\documentclass{article}

\PassOptionsToPackage{numbers}{natbib}

\usepackage{stfloats}

\usepackage[preprint]{neurips_2026}

\usepackage[utf8]{inputenc} 
\usepackage[T1]{fontenc}    
\usepackage{hyperref}       
\usepackage{url}            
\usepackage{booktabs}       
\usepackage{amsfonts}       
\usepackage{nicefrac}       
\usepackage{microtype}      
\usepackage[table]{xcolor}  
\usepackage{amssymb}        
\usepackage{amsmath}
\usepackage{amsthm}
\usepackage{siunitx}       
\usepackage{algorithm}
\usepackage{algpseudocode}
\usepackage{multirow}       
\usepackage{tabularray}
\usepackage{bm}             
\usepackage{graphicx}
\usepackage{wrapfig}
\usepackage{float}

\usepackage{xspace}
\usepackage{ulem}
\usepackage[most]{tcolorbox}

\definecolor{myblue}{HTML}{0F2A4A}   

\newtcolorbox[auto counter, number freestyle={\noexpand\arabic{\tcbcounter}}]{definedbox}[2][]{%
    enhanced,
    colback=myblue!6!white,
    colframe=myblue!85!white,
    title=Prompt~\thetcbcounter: #2,
    #1
}

\lstdefinelanguage{json}{
    showstringspaces=false,
    literate=
     *{0}{{{\color{purple}0}}}{1}
      {1}{{{\color{purple}1}}}{1}
      {2}{{{\color{purple}2}}}{1}
      {3}{{{\color{purple}3}}}{1}
      {4}{{{\color{purple}4}}}{1}
      {5}{{{\color{purple}5}}}{1}
      {6}{{{\color{purple}6}}}{1}
      {7}{{{\color{purple}7}}}{1}
      {8}{{{\color{purple}8}}}{1}
      {9}{{{\color{purple}9}}}{1}
      {:}{{{\color{black}{:}}}}{1}
      {,}{{{\color{black}{,}}}}{1}
      {\{}{{{\color{black}{\{}}}}{1}
      {\}}{{{\color{black}{\}}}}}{1}
      {[}{{{\color{black}{[}}}}{1}
      {]}{{{\color{black}{]}}}}{1},
    string=[s]{"}{"},
    stringstyle=\color{teal},
    comment=[l]{:\ "},
    morecomment=[l]{:"},
    commentstyle=\color{black},
}

\theoremstyle{plain}

\newtheorem{proposition}{Proposition}

\title{\textsc{FlowEdit}: Information-Theoretic Control of LLM Reasoning Flows for Ill-posed Problems Involving Conflicts}

\author{%
  Sizhe~Tang\\
  The George Washington University\\
  \texttt{s.tang1@gwu.edu} \\
  \And
  Guangyu~Jiang \\
  The George Washington University\\
  \texttt{guangyu.jiang@gwu.edu}\\
  \And
  Yu~Li\\
  The George Washington University\\
  \texttt{yul@gwu.edu}\\
  \And
  Rongqian~Chen\\
  The George Washington University\\
  \texttt{rongqianc@gwu.edu}\\
  \And
  Ioannis G. Kevrekidis\\
  Johns Hopkins University\\
  \texttt{yannisk@jhu.edu} \\
  \And
  Tian~Lan\\
  The George Washington University\\
  \texttt{tlan@gwu.edu}
}

\begin{document}

\maketitle

\begin{abstract}
Large Language Models (LLMs) have demonstrated strong performance on well-specified reasoning tasks where a feasible answer exists. However, problems encountered in the open world can become ill-posed due to inconsistent conditions, conflicting statements, or mutually incompatible requirements. It leads to ill-posed problems involving conflicts that admit no valid responses. We argue that effective reasoning of such ill-posed problems involving conflicts require novel LLM capabilities to make hidden conflicts explicit, maintain competing hypotheses via multiple reasoning branches when warranted, and generate alternative responses in a single pass, all of which are challenging due to the inherent limitation of next-token prediction mechanism in LLMs. To this end, we propose \textsc{FlowEdit}, a novel framework that leverages information-theoretic principles to actively quantify and regulate internal reasoning flows of LLMs, for generating a full set of alternative responses under valid hypotheses in a single pass. \textsc{FlowEdit} can be viewed as enforcing a branch-aware reasoning process using two dual information-theoretic objectives on the model's internal reasoning representations: maximizing the information flow from each selected hypothesis to the branch outcome, while minimizing the overlap and conditional dependence across sibling branches, to provide a diverse, informative set of responses with broad coverage. We show that this is achieved through tractable variational bounds under boundary embeddings being \(\epsilon\)-sufficient, optimizing the underlying conditional mutual information in LLM reasoning process. Extensive experiments demonstrate that \textsc{FlowEdit} outperforms leading proprietary models, improving exact-set-match accuracy by 68\%, while boosting overall response informativeness by 24\%. We further show that flow regulation surfaces in the token stream as a structured redistribution of next-token entropy that concentrates inside each branch, amplifies at the boundaries between flows, and scales with the number of flows the problem requires.
\end{abstract}

\begin{figure}[thb]
  \vspace{-0.1in}
  \centering
  \includegraphics[width=\linewidth]{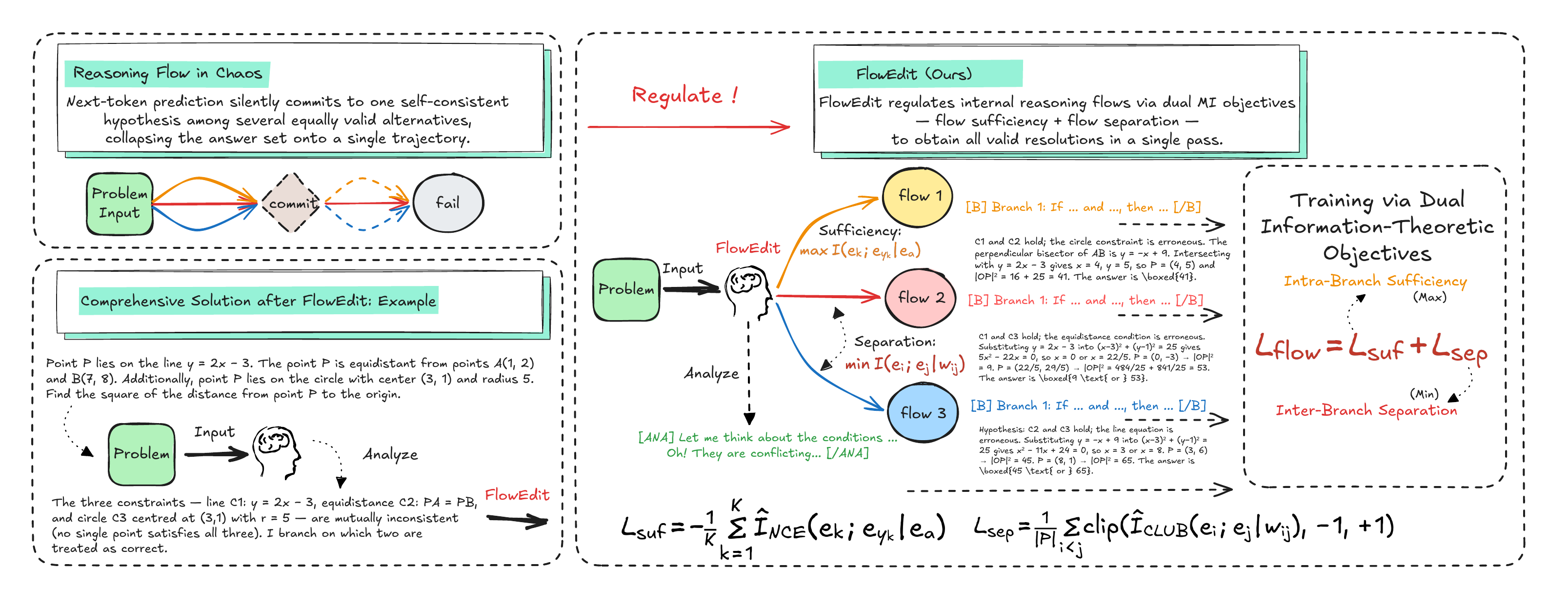}\hfill
  \vspace{-0.1in}
  \caption{Overview of \textsc{FlowEdit}. On ill-posed problems involving conflicts, next-token prediction silently commits to one self-consistent hypothesis and collapses the answer set onto a single trajectory (left). \textsc{FlowEdit} regulates the model's internal reasoning flows via two dual information-theoretic objectives on boundary representations (right). One example involving conflicts is demonstrated.}
  \vspace{-0.1in}
  \label{fig:overview}
\end{figure}

\section{Introduction}

Large language models (LLMs) have shown strong performance across diverse tasks \citep{tang2026agent, fang2026mint, jiang2025agentic, zhang2026metric}, with mathematical reasoning~\citep{cobbe2021training} standing out as a central capability. Current approaches primarily focus on well-specified problems where a feasible answer exists and train models to produce a single reasoning trace, which is supervised either by step-level rationales through chain-of-thought distillation~\citep{wei2022chain, yu2023metamath} or by a scalar reward on the final output through reinforcement learning with verifiable rewards~\citep{shao2024deepseekmath, guo2025deepseek}. This has proven to be effective on well-defined problems, where a single deterministic line of thought or logic suffices to reach the unique correct answer. However, reasoning problems encountered in the open world can often become ill-posed~\citep{zhao2024exploring, ma2026large,tian2025vcsearch,li2025questbench,xue2025reliablemath}. LLMs that lack the ability to analyze the inherent problem structures, examine implicit deficits/conflicts, or regulate how reasoning process unfolds demonstrate limited performance on ill-posed problems.

Among various forms of ill-posed problems, this paper considers ill-posed problems involving conflicts that admit no valid responses due to inconsistent conditions, conflicting states, or mutually incompatible requirements. Such type of ill-posed problems involving conflicts are routinely encountered in areas such as real-world decision-making, operations research, and systems engineering~\citep{OC1,OC2,OC3}. While recent work have studied ill-posed problems with missing conditions~\citep{zhao2024exploring, tian2025vcsearch} and underspecified reasoning~\citep{ma2026large,li2025questbench}, these are largely cast into a detection problem to flag or reject ill-posed inputs~\citep{xue2025reliablemath,rajpurkar2018know} or a post-hoc inference problem about response confidence \cite{kadavath2022language, kuhn2023semantic, lin2022teaching}. 

We show that for ill-posed problems involving conflicts, current LLMs either reject the input altogether \cite{tian2025vcsearch} or silently commit to one hypothesis (regarding the conflicts) among several equally possible alternatives \cite{ma2026large}. In contrast, our goal is to enable a more proactive reasoning behavior by making hidden conflicts explicit, analyzing and maintaining competing hypotheses via multiple reasoning branches when warranted, and generating alternative responses/recommendations in a single pass. This is challenging as LLMs' next-token prediction mechanism favors sequential elaboration of an initial commitment over the parallel maintenance of several \cite{bachmann2024pitfalls}. Novel solutions are needed to quantify and regulate multiple internal reasoning flows/branches in LLMs, to generate a full set of alternative responses under valid hypotheses in a single pass. 

We introduce \textsc{FlowEdit}, a training framework that operationalizes this notion of regulating internal reasoning flows and maintaining competing hypotheses through an information-theoretic control on a language model's internal representations and information flows. In particular, we identify the existence of conflicts and treat each reasoning flow as an object summarized by a designated hidden state along the generation trajectory, so that the abstract notion of enforcing multiple reasoning flows/branches becomes a concrete requirement on the geometry of these representations. A pair of dual objectives--expressed as conditional mutual information--make this requirement precise: each reasoning flow must retain the maximal information needed to maintain its own hypothesis and derive its own conclusion, while distinct flows should have minimal overlap beyond shared common problem analysis once the autoregressive context preceding them has been accounted for, as the reasoning process unfolds in a single pass.

More precisely, \textsc{FlowEdit} optimizes the max-min dual objectives by quantifying the conditional mutual information between selected LLM internal states. By maximizing the information
flow from each selected hypothesis to the branch outcome, while minimizing the conditional overlap across sibling branches, \textsc{FlowEdit} provides a diverse, informative set of responses with broad coverage in a single pass. We show that this optimization can be achieved through tractable variational bounds with boundary embeddings under an $\epsilon$-sufficient condition, provably optimizing the underlying conditional mutual information in LLM reasoning process. In particular, $\epsilon$-sufficiency makes the conditioning choice principled rather than heuristic: regulating cross-flow dependence against the analysis embedding, rather than a prompt-only embedding, provably isolates the residual redundancy we wish to penalize from the shared structure that distinct flows are entitled to encode.

\textsc{FlowEdit} demonstrates superior proactive reasoning behavior on ill-posed problems involving conflicts. Across three problem domains and ground-truth branch counts $K^\star\!\in\!\{1,2,3,4\}$, \textsc{FlowEdit}-Qwen3-4B-Base improves exact-set-match accuracy by 68\% and information recovery by 24\% over the strongest closed-source baseline, with gains widening as $K^\star$ grows; token-level analysis further shows these gains coincide with a structured redistribution of next-token entropy that concentrates inside each branch and amplifies at the boundaries between flows.

The primary contributions of this paper are as follows:
\begin{itemize}
  \item We reframe ill-posed mathematical reasoning with conflicting conditions as a flow-regulation problem, training models to enumerate the full set of valid resolutions in a single pass rather than detect and reject.
  \item We derive a flow-regulation objective from two conditional mutual information requirements, optimized through directionally aligned bounds under an $\epsilon$-sufficiency conditioning that provably isolates residual cross-flow dependence.
  \item We construct a dataset of problems involving conflicts spanning three domains and ground-truth branch counts $K^\star\!\in\!\{1,2,3,4\}$, with verified hypothesis–answer pairs per branch.
  \item We show \textsc{FlowEdit} outperforms the strongest closed-source baseline by 68\% on exact-set-match and 24\% on information recovery, and identify a token-level entropy signature—concentration inside branches, amplification at flow boundaries—that gives a representation-level account of the gains.
\end{itemize}

\section{Related Works and Background}
\label{sec:related}

\subsection{Mathematical Reasoning with Language Models}
\label{sec:related_math}
Research on mathematical reasoning with language models has progressed along three tracks: prompting-based methods that elicit multi-step reasoning from frozen models \citep{wei2022chain,wang2022self,yao2023tree,gao2023pal,chen2022program,li2026reason}, supervised fine-tuning that distills step-level traces into smaller models \citep{yu2023metamath,luo2023wizardmath,guan2025rstar}, and reinforcement learning with verifiable rewards against a single gold answer \citep{shao2024deepseekmath,guo2025deepseek, yu2026interactive}. All three share the same operative assumption: every problem is well-posed and admits a unique correct answer. We depart from this point-valued target and treat the answer to an ill-posed problem involving conflicts as a set of mutually exclusive resolutions, training the model to enumerate them within a single structured response.

\subsection{Ill-Posed Mathematical Reasoning and Abstention}
\label{sec:related_illposed}
Recent work examines LLM behavior on problems that violate the well-posed assumption, through benchmarks of logical traps, unreasonable inputs, and missing or contradictory conditions \citep{zhao2024exploring,ma2026large,tian2025vcsearch,li2025questbench}, alignment strategies that fine-tune models to output \texttt{unsolvable} \citep{xue2025reliablemath}, and parallel efforts in open-domain QA on unanswerable queries \citep{zhang2024r,rajpurkar2018know}. Despite their methodological variety, all of these threads frame ill-posed input as a detection problem: the model should identify the input as unsolvable and abstain, or solicit clarification to restore well-posedness. This framing discards a structure central to mathematical practice, namely that an ill-posed problem involving conflicts is typically not unsolvable but multiply solvable, with each self-consistent subset of its conditions defining a well-posed subproblem with a distinct answer. We depart from the detection-and-rejection paradigm and train LLMs to constructively enumerate the full set of valid resolutions; to our knowledge ours is the first work to formulate ill-posed mathematical reasoning involving conflicts as an information flow regulation task and to provide both a training objective and a dataset targeting this setting.

A parallel tradition in numerical optimization addresses ill-posed problems involving conflicts at the level of explicit constraints. The Maximum Feasible Subsystem problem seeks a largest cardinality subset of mutually inconsistent linear constraints that admits a joint solution \citep{amaldi1998approximability,chinneck2001fast}, and is approached at scale by RANSAC \citep{fischler1981random} and randomized Kaczmarz methods that exploit prior knowledge of conflict-free constraints \citep{strohmer2009randomized, lok2024subspace}. These methods operate on explicit constraint matrices and return a single resolution, whereas our setting requires the model to surface implicit conflicts from the problem statement and enumerate the full set of resolutions in a single pass.

\section{Methodology}
\subsection{Problem Formulation and Preliminaries}
\label{sec:problem-formulation}

We formalize the response to an ill-posed mathematical problem involving conflicts as a task in which a single model output must carry, and keep separate, several reasoning flows in parallel, and we specify the conditions under which a response is correct at the level of information rather than surface tokens. Well-posed problems are subsumed as the degenerate case $K^\star(x) = 1$, in which the framework reduces to standard single-trace reasoning; the formulation is therefore a strict generalization of the point-valued target adopted in prior work, not a separate regime. 

\paragraph{Ill-posed problems and reasoning flows.} 
Let $x$ denote a problem statement and $\mathcal{C}(x) = \{c_1, \ldots, c_m\}$ the conditions stated in $x$. We call $x$ \emph{ill-posed} if no assignment of the problem variables satisfies every $c_i$ simultaneously, although strict subsets of $\mathcal{C}(x)$ are jointly satisfiable. A \emph{hypothesis} $z \subseteq \mathcal{C}(x)$ is a self-consistent subset of conditions, and $\mathcal{Z}^{\star}(x)$ collects those subsets that are maximal under inclusion among self-consistent ones and induce a well-posed subproblem with a unique solution. With $K^{\star}(x) = |\mathcal{Z}^{\star}(x)|$ and $y_k = g(x, z_k)$ the answer under the domain solution map $g$, the ground-truth target is
\begin{equation}
\mathcal{Y}^{\star}(x) = \{g(x, z_k) : z_k \in \mathcal{Z}^{\star}(x)\} = \{y_1, \ldots, y_{K^{\star}(x)}\},
\end{equation}
the well-posed case $K^{\star}(x) = 1$ retained as a degenerate singleton. Each $z_k$ defines a distinct deterministic \emph{reasoning flow} from $x$ to $y_k$; ill-posedness is the regime in which $K^{\star}(x) > 1$ and a correct response must carry several such flows in parallel within a single output. Our methodological goal is to regulate these flows: to ensure that they form inside the model and remain mutually distinct, rather than collapsing onto whichever flow next-token prediction commits to first.

\paragraph{Where reasoning flows become observable.}
Regulating reasoning flow at the representational level requires a place to address each flow individually. We obtain this addressability with the smallest structural intervention sufficient for the purpose: the policy $\pi_\theta$ is trained to produce an output that interleaves an analysis segment $a$ with a sequence of \emph{branch blocks}, one per flow,
\begin{equation}
o = \texttt{[ANA]}\, a\, \texttt{[/ANA]}\; \texttt{[B]}\, \hat{z}_1\, \texttt{[SEP]}\, \hat{y}_1\, \texttt{[/B]}\, \cdots\, \texttt{[B]}\, \hat{z}_{\hat{K}}\, \texttt{[SEP]}\, \hat{y}_{\hat{K}}\, \texttt{[/B]}\; \texttt{[END]},
\label{eq:structured-output}
\end{equation}
where $\hat{K}$ is chosen by the model through emission of \texttt{[END]} rather than supplied externally, and the structural tokens carry no semantic content of their own. The role of \eqref{eq:structured-output} is not to prescribe a format but to designate hidden states at which each flow is summarized: writing $h_t(\cdot)$ for the causal-LM hidden state at token $t$, we read off
\[
e_x = h_{\texttt{[ANA]}}(x), \qquad e_a = h_{\texttt{[/ANA]}}(x, a), \qquad e_k = h_{\texttt{[SEP]}_k}(\cdots), \qquad e_{y_k} = h_{\texttt{[/B]}_k}(\cdots) \;\in\; \mathbb{R}^d
\]
as the prompt summary, the analysis summary, the $k$-th flow's hypothesis representation, and its answer representation \cite{tenney2019bert,hewitt2019structural}.

\subsection{Information-Theoretic Reasoning Flow Regulation}
\label{sec:flow-regulation}

A reasoning flow, as defined in Section \ref{sec:problem-formulation}, is determined by three boundary representations: the analysis $e_a$, a hypothesis $e_k$, and an answer $e_{y_k}$. Two properties of these representations distinguish a model that genuinely maintains several flows in parallel from one that emits multiple branches at the surface but reasons about them in a degenerate way, and they jointly motivate the developed loss.

The first property concerns each flow individually. A flow is well-formed only if its hypothesis representation $e_k$ carries, given the analysis $e_a$, the information from which its answer $e_{y_k}$ is determined; otherwise the branch-block is decorative, with the answer produced from $e_a$ alone or from cues unrelated to the committed hypothesis. We call this property \emph{flow sufficiency}, and it imposes on the loss a term that increases $I(e_k; e_{y_k} \mid e_a)$, pushing the hypothesis representation to carry the residual information that distinguishes its own subproblem from the analysis-level baseline.

The second property concerns the relationship between flows. Several flows coexist in a single output without collapse only if distinct hypothesis representations are not paraphrases of one another at the level of information: the more $e_i$ and $e_j$ share beyond what is licensed by the shared analysis and the autoregressive context that precedes them, the more the second flow is reducible to the first. We call this property \emph{flow separation}, and it imposes on the loss a term that reduces $I(e_i; e_j \mid w_{ij})$ as much as the optimization can afford, where $w_{ij}$ is a conditioning vector that absorbs the dependence we cannot and should not eliminate.

The remainder of this section constructs the flow-regulation loss $\mathcal{L}_{\mathrm{flow}}$ that enforces these two properties on the boundary embeddings of $\pi_\theta$. We work with $L_2$-normalized embeddings, $\tilde{e} = e/\|e\|_2$, suppressing the tilde, so that all representations lie on the unit sphere and the estimators below operate on a bounded, comparable scale. The overview of \textsc{FlowEdit} is demonstrated in Fig. \ref{fig:overview}.

\paragraph{Bounds aligned with the optimization directions.}
The two flow properties translate into the formal conditions
\begin{equation}
\max_{\theta}\ I(e_k\,;\,e_{y_k} \mid e_a)\ \ \text{for each } k, \qquad \min_{\theta}\ I(e_i\,;\,e_j \mid w_{ij})\ \ \text{for each pair } i<j,
\label{eq:two-conditions}
\end{equation}
which act on the same kind of quantity but ask the optimizer to move it in opposite directions. This asymmetry is consequential: maximizing a tractable lower bound provably increases the underlying MI, but minimizing the same lower bound does not reduce it---the bound can become loose without the dependence actually decreasing. Genuine reduction of MI requires an upper bound that majorizes the true quantity from above. We therefore choose two estimators by their bounding direction, InfoNCE~\cite{oord2018representation} for sufficiency and CLUB~\cite{cheng2020club} for separation:
\begin{equation}
\widehat{I}_{\mathrm{NCE}}(e_k\,;\,e_{y_k} \mid e_a) = \mathbb{E}\!\left[\log \frac{\exp\!\big(f_\phi(e_k, e_{y_k}, e_a)/\tau\big)}{\tfrac{1}{N}\sum_{n=1}^{N}\exp\!\big(f_\phi(e_k, e_{y_k}^{(n)}, e_a)/\tau\big)} \right],
\label{eq:nce}
\end{equation}
\begin{equation}
\widehat{I}_{\mathrm{CLUB}}(e_i\,;\,e_j \mid w_{ij}) = \mathbb{E}_{p(e_i, e_j, w_{ij})}\!\big[\log q_\psi(e_j \mid e_i, w_{ij})\big] - \mathbb{E}_{p(e_i, w_{ij})\,p(e_j \mid w_{ij})}\!\big[\log q_\psi(e_j \mid e_i, w_{ij})\big],
\label{eq:club}
\end{equation}
with $f_\phi$ a learned scoring function, $\tau = 0.07$, negatives $\{e_{y_k}^{(n)}\}$ drawn from other branches in the minibatch, and the second CLUB expectation taken over shuffled $(e_j, w_{ij})$ pairs. The variational network $q_\psi(v \mid u, w)$, parameterized as a diagonal-Gaussian MLP that predicts the mean and log-variance of $e_j$ given $(e_i, w_{ij})$, approximates the true conditional distribution $p(e_j \mid e_i, w_{ij})$; the CLUB difference is a valid upper bound on $I(e_i; e_j \mid w_{ij})$ to the extent that this approximation is accurate, and $q_\psi$ is therefore updated alongside $\theta$ during training. CLUB is preferred over geometric penalties such as cosine or orthogonality, which capture only second-order dependence and cannot register the higher-order structure two reasoning flows can share.

\paragraph{Conditioning that isolates the flow-level residual.}
The estimators in~\eqref{eq:nce}--\eqref{eq:club} are only as faithful as their conditioning. The conditioning vector $w_{ij}$ in the separation term defines what counts as redundancy: it must remove the dependence that two flows are entitled to share, so that the regularizer acts on residual cross-flow information rather than on a baseline guaranteed by construction, but it must not remove the dependence that distinguishes them, lest the regularizer be driven to zero on precisely the multi-flow inputs where it should be active. The second of these requirements singles out the analysis representation $e_a$ from natural alternatives such as a prompt-only embedding $e_x = h_{\texttt{[ANA]}}(x)$, a preference we now formalize. Define $I^\star_{ij} := I(E_i; E_j \mid X, A)$, the flow-level residual the
regularizer is meant to penalize, and $\Delta_{ij} := I(E_i; E_j \mid X) -
I^\star_{ij}$, the share of cross-flow dependence contributed by $A$ beyond
$X$. By construction $\Delta_{ij} \ge 0$ on multi-flow problems---each
$a^\star$ verbalises a partition that determines every branch, contributing
dependence beyond what $X$ alone supplies, with $\Delta_{ij} > 0$ in the
regime where $\mathcal{L}_{\mathrm{sep}}$ is intended to act.

\begin{proposition}
\label{prop:e_a-conditioning}
Suppose $I((X,A); (E_i,E_j) \mid e_a) \le \epsilon_a$ and $I(X; (E_i,E_j) \mid e_x) \le \epsilon_x$, i.e., the boundary embeddings are $\epsilon$-sufficient summaries of their attention contexts. Then
\begin{align}
\big|\,I(E_i; E_j \mid e_a) - I^{\star}_{ij}\,\big| &\;\le\; \epsilon_a, \label{eq:prop-ea-bound}\\
\big|\,I(E_i; E_j \mid e_x) - (I^{\star}_{ij} + \Delta_{ij})\,\big| &\;\le\; \epsilon_x, \label{eq:prop-ex-bound}\\
I(E_i; E_j \mid e_x) - I(E_i; E_j \mid e_a) &\;\ge\; \Delta_{ij} - (\epsilon_a + \epsilon_x). \label{eq:prop-gap}
\end{align}
The same lower bound~\ref{eq:prop-gap} holds when $e_x$ is replaced by any $\sigma(X)$-measurable, $\epsilon_x$-sufficient statistic of $X$.
\end{proposition}

The proposition partitions cross-flow dependence into a residual we wish to penalize and a component $\Delta_{ij}$ contributed by the analysis. Conditioning on $e_a$ targets only the former; conditioning on $e_x$ pulls both terms down jointly, which means the gradient delivered to $\theta$ contains a component that pushes the boundary embeddings toward forgetting the partition $A$ verbalised. On multi-flow inputs, where the partition is exactly the structure several flows must encode differently, this gradient component drives the very flow collapse the regularizer is meant to prevent. The proof is given in Appendix \ref{app:proof} and the no-analysis ablation is given in Section Experiment \ref{sec:ablation}.

With $e_a$ fixed as the analysis-level baseline, the conditioning vector must absorb one further source of dependence specific to the autoregressive setting. In any causal LM, the receptive field of $e_j$ for $j > i$ contains $e_i$ via attention over the intervening tokens, so $I(e_i; e_j \mid e_a)$ has a baseline level fixed by architecture rather than by content; minimizing it without absorbing this baseline penalises the model for an information-flow it cannot avoid. We therefore append a preceding-flow summary $\bar{e}_{<m} = \tfrac{1}{m-1}\sum_{j<m} e_j$ with $\bar{e}_{<1} := 0$, giving
\begin{equation}
w_{ij} \;=\; \big[\,e_a\,;\;\bar{e}_{<\min(i,j)}\,\big],
\label{eq:wij}
\end{equation}
under which $\widehat{I}_{\mathrm{CLUB}}$ measures only the residual cross-flow redundancy. The sufficiency term \eqref{eq:two-conditions} requires a strict subset of these absorptions: only $e_a$ matters, since the autoregressive baseline is between sibling hypothesis representations and not between a hypothesis and its own answer. We accordingly set $w = e_a$ in $\widehat{I}_{\mathrm{NCE}}$, recovering~\eqref{eq:nce} as written.

\paragraph{End-to-End Training Procedure.} With the directional bounds of \eqref{eq:nce}--\eqref{eq:club} and the conditioning of \eqref{eq:wij} in place, the per-property losses
\begin{align}
\mathcal{L}_{\mathrm{suf}}(\theta) &= -\frac{1}{K}\sum_{k=1}^{K} \widehat{I}_{\mathrm{NCE}}(e_k\,;\,e_{y_k} \mid e_a), \label{eq:lsuf}\\
\mathcal{L}_{\mathrm{sep}}(\theta) &= \frac{1}{|\mathcal{P}|}\sum_{(i,j)\in\mathcal{P}} \mathrm{clip}\!\big(\widehat{I}_{\mathrm{CLUB}}(e_i\,;\,e_j \mid w_{ij}),\,-1,\,+1\big), \label{eq:lsepterm}
\end{align}
with $\mathcal{P} = \{(i,j) : 1 \le i < j \le K\}$, combine into the flow-regulation loss
\begin{equation}
\mathcal{L}_{\mathrm{flow}}(\theta) \;=\; \mathcal{L}_{\mathrm{suf}}(\theta) \;+\;\mathcal{L}_{\mathrm{sep}}(\theta).
\label{eq:lflow}
\end{equation}
The per-pair clip in $\mathcal{L}_{\mathrm{sep}}$ bounds the contribution of any single pair-batch to a fixed nats budget; without it, the variance of $\widehat{I}_{\mathrm{CLUB}}$ at the small per-step pair count $|\mathcal{P}|$ available within one minibatch admits occasional excursions whose magnitude is uncorrelated with the underlying redundancy and whose effect on $\theta$ is therefore noise. To provide a substrate of structured generation on which $\mathcal{L}_{\mathrm{flow}}$ can act, we pair it with the standard token-level generation loss over the structured output \eqref{eq:structured-output},
\begin{equation}
\mathcal{L}_{\mathrm{gen}}(\theta) \;=\; -\log p_\theta(a^{\star} \mid x) - \sum_{k=1}^{K^{\star}} \big[\log p_\theta(z_k^{\star} \mid x, a^{\star}, z_{<k}^{\star}, y_{<k}^{\star}) + \log p_\theta(y_k^{\star} \mid x, a^{\star}, z_{\le k}^{\star}, y_{<k}^{\star})\big],
\label{eq:lgen}
\end{equation}
and the policy is trained on $\mathcal{L}_{\mathrm{gen}} + \lambda \,\mathcal{L}_{\mathrm{flow}}$ where \(\lambda\) controls the relative weight.

\section{Experiments}
\label{sec:experiments}

\definecolor{avgshade}{gray}{0.92}
\newcommand{\Lsep}{\mathcal{L}_{\mathrm{sep}}}
\definecolor{ourshade}{HTML}{D6E6F2}

\subsection{Experiment Setup}
\label{sec:setup}

\textbf{Dataset.} We construct a dataset of $5{,}000$ ill-posed reasoning problems involving conflicts. Each instance is a tuple $(x, a, \{(z_k, y_k)\}_{k=1}^{K^\star})$: a problem $x$, an analysis $a$ identifying its underspecified conditions, and $K^\star\!\in\!\{1,2,3,4\}$ valid resolutions, each pairing a hypothesis $z_k$ with the answer $y_k$ that follows under it. We explicitly include well-posed instances ($K^\star\!=\!1$) so the model cannot exploit a fixed enumeration prior. Problems span three domain categories (Pure Math, Daily, Application) and three difficulty tiers. Each instance is rendered into a single causal sequence $\texttt{[ANA]}\,a\,\texttt{[/ANA]}\,\texttt{[B]}\,z_1\,\texttt{[SEP]}\,y_1\,\texttt{[/B]}\cdots\texttt{[END]}$, exposing one boundary token per hypothesis and answer on which $\mathcal{L}_{\mathrm{flow}}$ is computed. Further details are in Appendix~\ref{appn:dataset}.

\textbf{Baselines and training.} We compare against closed-source prompted models (Claude Haiku 4.5, GPT-5, Gemini 2.5 Pro) and the same open-source backbones under prompting only, each evaluated under a \emph{blind} prompt (no mention of ill-posedness) and an \emph{ill-posed} prompt (explicit enumeration instruction); the ill-posed prompt is held identical across baselines to isolate the training contribution from prompt engineering. We instantiate \textsc{FlowEdit} on Qwen3-4B-Base and Qwen2.5-3B-Instruct \cite{yang2025qwen3}, training both under $\mathcal{L}_{\mathrm{gen}} + \lambda\ \mathcal{L}_{\mathrm{flow}}$ for 2{,}000 optimizer steps with \(\lambda=0.1\), using the \emph{ill-posed} prompt. The parameter sensitivity experiments are illustrated in Appendix \ref{appn:parameter_sensitivity} We report exact-set-match (EM) and information recovery (IR), broken down by domain and by $K^\star \in \{1,2,3,4\}$.

\subsection{Main Results}
\label{sec:main_results}

\begin{figure}[H]
  \vspace{-0.1in}
  \centering
  \includegraphics[width=0.5\linewidth]{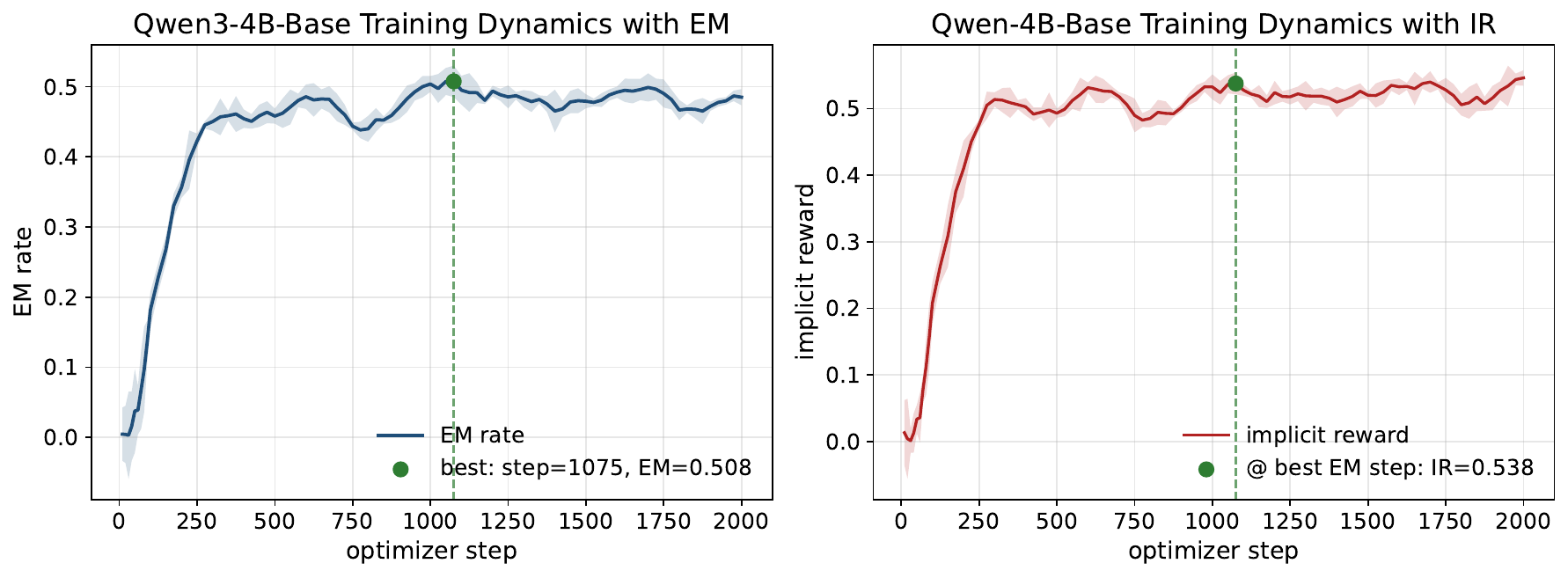}\hfill
  \includegraphics[width=0.5\linewidth]{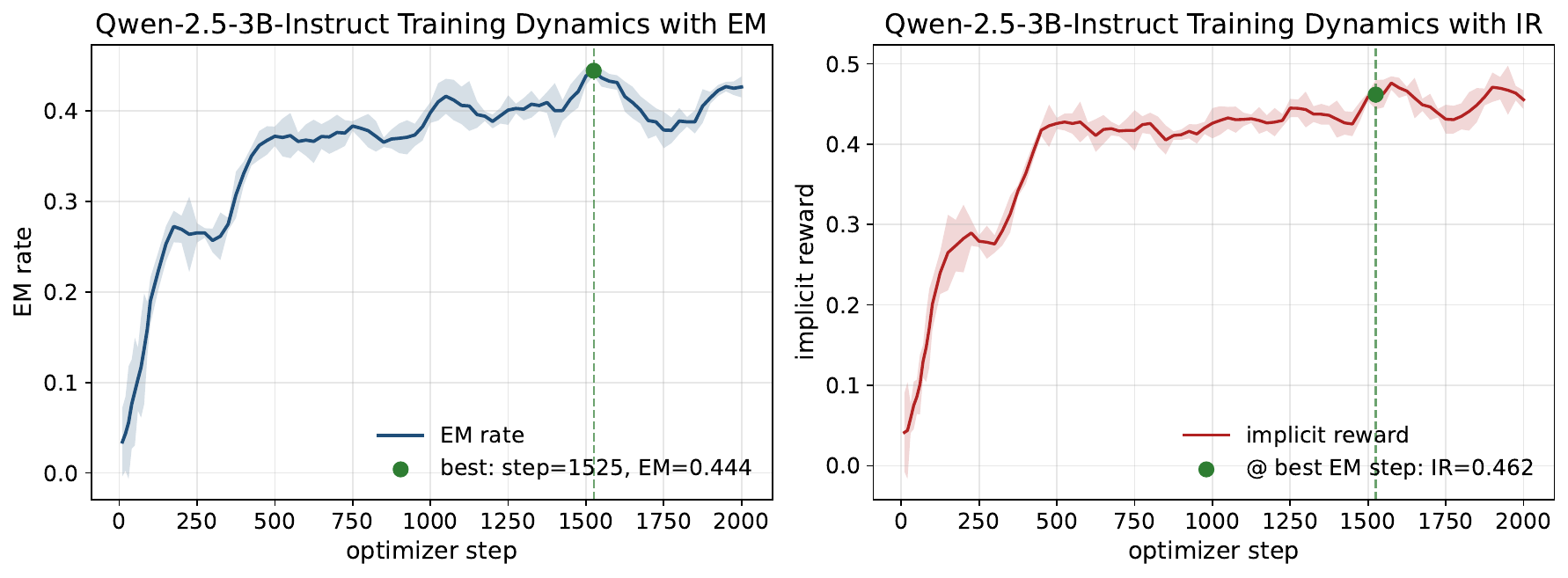}\hfill
  \vspace{-0.1in}
  \caption{Training dynamics of \textsc{FlowEdit} based on Qwen-3-4B Base and Qwen-2.5-3B-Instruct with metrics of EM and IR. It is illustrated that the training of \textsc{FlowEdit} can converge stably with high performance. Shaded regions indicate standard deviation across three seeds.}
  \vspace{-0.1in}
  \label{fig:training_dynamics}
\end{figure}

\emph{Exact-set-match} (EM) credits a response only when $\hat{\mathcal{Y}}(x) = \mathcal{Y}^{\star}(x)$, treating every deviation as a uniform failure and remaining silent on the share of valid resolutions a partially correct response recovers. We complement it with \emph{information recovery} (IR),
\begin{equation}
  \mathrm{IR}(x) \;=\;
  \frac{|\hat{\mathcal{Y}}(x) \cap \mathcal{Y}^{\star}(x)|}
       {\max\!\big(|\hat{\mathcal{Y}}(x)|,\, |\mathcal{Y}^{\star}(x)|\big)},
  \label{eq:ir}
\end{equation}
which penalises missing flows through the numerator (a silent commitment on $K^{\star}{=}4$ yields $\mathrm{IR}\!\leq\!1/4$) and spurious flows through the denominator, ruling out the trivial strategy of over-emitting to maximise coverage. IR reduces to EM on $K^{\star}{=}1$ inputs.

\paragraph{Training Dynamics.} Figure~\ref{fig:training_dynamics} tracks EM and IR across training. The two metrics target the failure modes $\mathcal{L}_{\mathrm{flow}}$ prevents: missing flows depress IR, while flow collapse onto duplicate hypotheses depresses EM without lowering IR. At both scales the two curves ascend jointly and peak in the same region ($\mathrm{EM}{=}0.508 / \mathrm{IR}{=}0.538$ at 4B, step ${\sim}1075$; $\mathrm{EM}{=}0.444 / \mathrm{IR}{=}0.462$ at 3B, step ${\sim}1525$). Joint ascent is the trajectory-level counterpart of the dual-objective design: $\mathcal{L}_{\mathrm{suf}}$ raises IR by anchoring each branch to its answer, while $\mathcal{L}_{\mathrm{sep}}$ raises EM by enforcing set-distinctness. Decoupling---IR rising without EM---does not appear at either scale, indicating that the two terms are jointly active rather than competing for capacity. We adopt single-criterion selection on validation EM for all results in Table~\ref{tab:main}.

\paragraph{Scaling behaviour.} The 4B and 3B trajectories share their qualitative shape, indicating that $\mathcal{L}_{\mathrm{flow}}$ transfers across scale without re-tuning. Scaling from 3B to 4B improves IR by $0.076$ and EM by $0.064$ at comparable rates, consistent with the dual objective being jointly capacity-bound rather than one term saturating first. The persistent gap of ${\sim}0.03$ between IR and EM at both scales reflects the structural asymmetry of the metrics: set-equality is strictly stronger than coverage at every capacity.

\paragraph{Performance and informativeness.} \textsc{FlowEdit}-Qwen3-4B-Base attains $\mathrm{EM}{=}0.47$ and $\mathrm{IR}{=}0.51$, exceeding the strongest closed-source baseline (Haiku-4.5 with the ill-posed prompt, $0.28 / 0.41$) by $+68\%$ on EM and $+24\%$ on IR, consistently across all three domains and at both scales. The smaller relative gain on IR is diagnostic: baselines that silently commit on $K^{\star}{>}1$ inputs already accrue $1/K^{\star}$ partial credit, so IR headroom is structurally narrower than EM. Against open-source prompted baselines the gap widens to $+88\%$ EM and $+24\%$ IR (over Qwen3-4B with the ill-posed prompt), confirming the gain stems from training rather than base-model capacity. The advantage is most pronounced where silent commitment is most costly: on $K^{\star}{=}2$ inputs, \textsc{FlowEdit} reaches $\mathrm{EM}{=}0.40, 0.54, 0.65$ across domains versus $0.20, 0.27, 0.52$ for the strongest closed-source baseline---the regime where parallel hypothesis maintenance pays off most against the $1/K^{\star}$ ceiling of single-hypothesis prompting.

\definecolor{bannerpink}{RGB}{245,215,215}
\definecolor{sectionshade}{gray}{0.88}
\begin{table}[t]
\centering
\caption{Branch-level performance and informativeness on three problem
domains (\textit{Pure Math}, \textit{Daily}, \textit{Application}),
broken down by ground-truth branch count
$K^{\star}\!\in\!\{1,2,3,4\}$, with per-domain and overall weighted averages. Best result in each column is in \textbf{bold}. \textsc{FlowEdit} consistently outperforms across all baselines.}
\label{tab:main}
\setlength{\tabcolsep}{4pt}
\renewcommand{\arraystretch}{1.10}
\resizebox{\linewidth}{!}{%
\begin{tabular}{l c ccccc ccccc ccccc c}
\toprule

\rowcolor{sectionshade}
\multicolumn{18}{l}{\textbf{Performance (Exact-Set-Match)}} \\

\multirow{2}{*}{\textbf{Method}} & \multirow{2}{*}{\textbf{Model}}
  & \multicolumn{5}{c}{\textbf{Pure Math}}
  & \multicolumn{5}{c}{\textbf{Daily}}
  & \multicolumn{5}{c}{\textbf{Application}}
  & \multirow{2}{*}{\textbf{Avg}} \\
\cmidrule(lr){3-7}\cmidrule(lr){8-12}\cmidrule(lr){13-17}
 & & $K{=}1$ & $K{=}2$ & $K{=}3$ & $K{=}4$ & \textbf{Avg}
   & $K{=}1$ & $K{=}2$ & $K{=}3$ & $K{=}4$ & \textbf{Avg}
   & $K{=}1$ & $K{=}2$ & $K{=}3$ & $K{=}4$ & \textbf{Avg}
   & \\
\midrule
\multicolumn{18}{l}{\textit{Prompting (closed-source)}} \\
Prompt (blind)    & Haiku 4.5      & 0.17 & 0.10 & 0.56 & 0.26 & 0.19 & 0.12 & 0.20 & 0.24 & 0.29 & 0.18 & 0.20 & 0.48 & 0.54 & 0.39 & 0.39 & 0.24 \\
Prompt (illposed) & Haiku 4.5      & 0.19 & 0.20 & 0.78 & 0.36 & 0.25 & 0.11 & 0.27 & 0.27 & 0.17 & 0.20 & 0.32 & 0.52 & 0.58 & 0.45 & 0.46 & 0.28 \\
Prompt (blind)    & GPT-5          & 0.19 & 0.20 & 0.44 & 0.24 & 0.22 & 0.27 & 0.23 & 0.16 & 0.14 & 0.23 & 0.25 & 0.46 & 0.58 & 0.39 & 0.40 & 0.27 \\
Prompt (illposed) & GPT-5          & 0.19 & 0.20 & \textbf{0.89} & 0.43 & 0.26 & 0.17 & 0.24 & 0.27 & 0.29 & 0.22 & 0.18 & 0.44 & 0.58 & 0.40 & 0.37 & 0.27 \\
Prompt (blind)    & Gemini 2.5 Pro & 0.10 & 0.20 & 0.22 & 0.16 & 0.13 & 0.12 & 0.26 & 0.18 & 0.43 & 0.19 & 0.18 & 0.42 & 0.38 & 0.27 & 0.32 & 0.21 \\
Prompt (illposed) & Gemini 2.5 Pro & 0.06 & 0.20 & 0.44 & 0.22 & 0.13 & 0.06 & 0.24 & 0.27 & \textbf{0.57} & 0.19 & 0.16 & 0.46 & 0.54 & 0.47 & 0.37 & 0.22 \\
\midrule
\multicolumn{18}{l}{\textit{Fine-tuning (open-source)}} \\
Prompt (blind)    & Qwen2.5-3B-Instruct & 0.03 & 0.05 & 0.00 & 0.02 & 0.03 & 0.04 & 0.02 & 0.02 & 0.00 & 0.03 & 0.02 & 0.02 & 0.00 & 0.03 & 0.02 & 0.03 \\
Prompt (illposed) & Qwen2.5-3B-Instruct & 0.03 & 0.05 & 0.22 & 0.03 & 0.05 & 0.01 & 0.06 & 0.08 & 0.00 & 0.04 & 0.00 & 0.02 & 0.04 & 0.04 & 0.02 & 0.04 \\
\rowcolor{ourshade}
\textsc{FlowEdit} & Qwen2.5-3B-Instruct & \textbf{0.43} & 0.25 & 0.33 & 0.13 & 0.37 & \textbf{0.33} & 0.42 & 0.41 & 0.14 & 0.37 & \textbf{0.41} & 0.40 & 0.67 & 0.18 & 0.44 & 0.39 \\
Prompt (blind)    & Qwen3-4B-Base            & 0.18 & 0.05 & 0.22 & 0.54 & 0.18 & 0.19 & 0.23 & 0.25 & \textbf{0.57} & 0.23 & 0.14 & 0.35 & 0.50 & \textbf{0.61} & 0.32 & 0.24 \\
Prompt (illposed) & Qwen3-4B-Base            & 0.09 & 0.10 & 0.33 & \textbf{0.55} & 0.14 & 0.13 & 0.32 & 0.37 & \textbf{0.57} & 0.26 & 0.16 & 0.31 & 0.54 & 0.60 & 0.32 & 0.25 \\
\rowcolor{ourshade}
\textsc{FlowEdit} & Qwen3-4B-Base       & 0.33 & \textbf{0.40} & 0.67 & \textbf{0.55} & \textbf{0.38} & 0.27 & \textbf{0.54} & \textbf{0.61} & \textbf{0.57} & \textbf{0.45} & \textbf{0.41} & \textbf{0.65} & \textbf{0.79} & 0.60 & \textbf{0.59} & \textbf{0.47} \\

\midrule
\rowcolor{sectionshade}
\multicolumn{18}{l}{\textbf{Informativeness (Information Recovery)}} \\

\multirow{2}{*}{\textbf{Method}} & \multirow{2}{*}{\textbf{Model}}
  & \multicolumn{5}{c}{\textbf{Pure Math}}
  & \multicolumn{5}{c}{\textbf{Daily}}
  & \multicolumn{5}{c}{\textbf{Application}}
  & \multirow{2}{*}{\textbf{Avg}} \\
\cmidrule(lr){3-7}\cmidrule(lr){8-12}\cmidrule(lr){13-17}
 & & $K{=}1$ & $K{=}2$ & $K{=}3$ & $K{=}4$ & \textbf{Avg}
   & $K{=}1$ & $K{=}2$ & $K{=}3$ & $K{=}4$ & \textbf{Avg}
   & $K{=}1$ & $K{=}2$ & $K{=}3$ & $K{=}4$ & \textbf{Avg}
   & \\
\midrule
\multicolumn{18}{l}{\textit{Prompting (closed-source)}} \\
Prompt (blind)    & Haiku 4.5      & 0.25 & 0.40 & 0.74 & 0.34 & 0.32 & 0.26 & 0.38 & 0.38 & 0.32 & 0.33 & 0.41 & 0.63 & 0.61 & 0.44 & 0.54 & 0.38 \\
Prompt (illposed) & Haiku 4.5      & 0.32 & 0.39 & 0.85 & 0.39 & 0.38 & 0.24 & 0.41 & 0.38 & 0.40 & 0.33 & \textbf{0.55} & 0.66 & 0.67 & 0.51 & 0.61 & 0.41 \\
Prompt (blind)    & GPT-5          & 0.27 & 0.37 & 0.52 & 0.27 & 0.31 & \textbf{0.33} & 0.35 & 0.39 & 0.39 & 0.35 & 0.34 & 0.55 & 0.65 & 0.44 & 0.49 & 0.38 \\
Prompt (illposed) & GPT-5          & 0.34 & \textbf{0.51} & \textbf{0.94} & 0.45 & \textbf{0.42} & 0.32 & 0.37 & 0.40 & 0.43 & 0.36 & 0.35 & 0.51 & 0.64 & 0.44 & 0.48 & 0.41 \\
Prompt (blind)    & Gemini 2.5 Pro & 0.21 & 0.38 & 0.43 & 0.25 & 0.26 & 0.25 & 0.33 & 0.25 & 0.57 & 0.29 & 0.38 & 0.55 & 0.49 & 0.35 & 0.46 & 0.33 \\
Prompt (illposed) & Gemini 2.5 Pro & 0.22 & 0.40 & 0.56 & 0.27 & 0.28 & 0.21 & 0.35 & 0.34 & 0.68 & 0.30 & 0.35 & 0.56 & 0.58 & 0.50 & 0.48 & 0.34 \\
\midrule
\multicolumn{18}{l}{\textit{Fine-tuning (open-source)}} \\
Prompt (blind)    & Qwen2.5-3B-Instruct & 0.05 & 0.05 & 0.07 & 0.05 & 0.05 & 0.08 & 0.11 & 0.12 & 0.07 & 0.10 & 0.04 & 0.08 & 0.03 & 0.10 & 0.06 & 0.08 \\
Prompt (illposed) & Qwen2.5-3B-Instruct & 0.07 & 0.08 & 0.30 & 0.09 & 0.09 & 0.05 & 0.15 & 0.16 & 0.11 & 0.11 & 0.06 & 0.05 & 0.11 & 0.13 & 0.07 & 0.09 \\
\rowcolor{ourshade}
\textsc{FlowEdit} & Qwen2.5-3B-Instruct & \textbf{0.43} & 0.25 & 0.41 & 0.37 & 0.39 & \textbf{0.33} & 0.48 & 0.50 & 0.39 & 0.42 & 0.41 & 0.45 & 0.71 & 0.42 & 0.48 & 0.43 \\
Prompt (blind)    & Qwen3-4B-Base            & 0.27 & 0.12 & 0.30 & \textbf{0.68} & 0.27 & 0.28 & 0.40 & 0.48 & \textbf{0.71} & 0.38 & 0.23 & 0.43 & 0.57 & \textbf{0.75} & 0.41 & 0.36 \\
Prompt (illposed) & Qwen3-4B-Base            & 0.22 & 0.31 & 0.52 & 0.65 & 0.28 & \textbf{0.33} & 0.49 & 0.58 & 0.68 & 0.45 & 0.29 & 0.48 & 0.56 & 0.72 & 0.44 & 0.41 \\
\rowcolor{ourshade}
\textsc{FlowEdit} & Qwen3-4B-Base       & 0.33 & 0.43 & 0.67 & \textbf{0.68} & 0.39 & 0.27 & \textbf{0.61} & \textbf{0.71} & \textbf{0.71} & \textbf{0.50} & 0.41 & \textbf{0.71} & \textbf{0.82} & \textbf{0.75} & \textbf{0.63} & \textbf{0.51} \\

\bottomrule
\end{tabular}}
\end{table}

\subsection{Token-Level Entropy Signatures of Flow Regulation}
\label{sec:entropy_analysis}

\begin{figure}[H]
  \vspace{-0.1in}
  \centering
  \includegraphics[width=\linewidth]{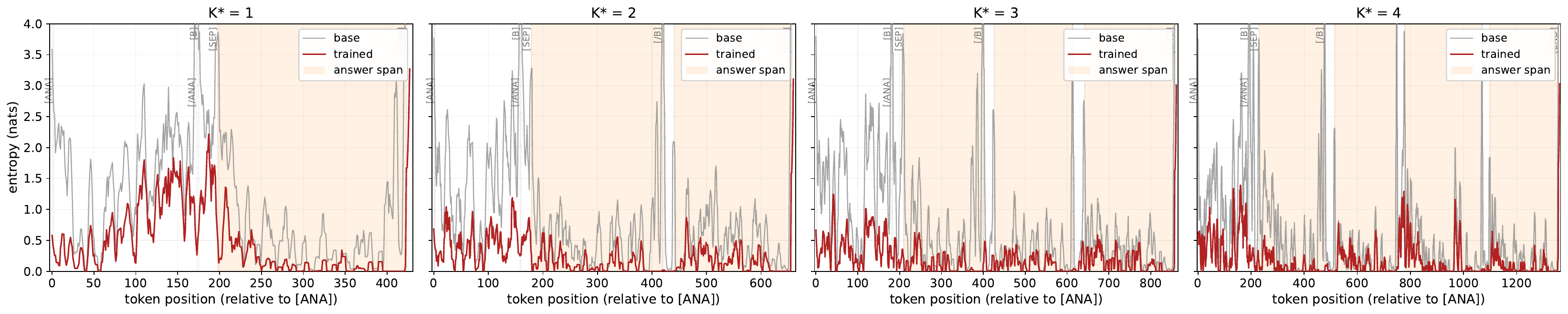}
  \caption{Per-sample teacher-forced next-token entropy, one example per $K^\star\in\{1,2,3,4\}$. The trained trace suppresses entropy inside answer spans and amplifies it at branch-opening \texttt{[B]} tokens.}
  \vspace{-0.1in}
  \label{fig:per_sample_entropy}
\end{figure}

\begin{figure}[H]
  \vspace{-0.1in}
  \centering
  \includegraphics[width=\linewidth]{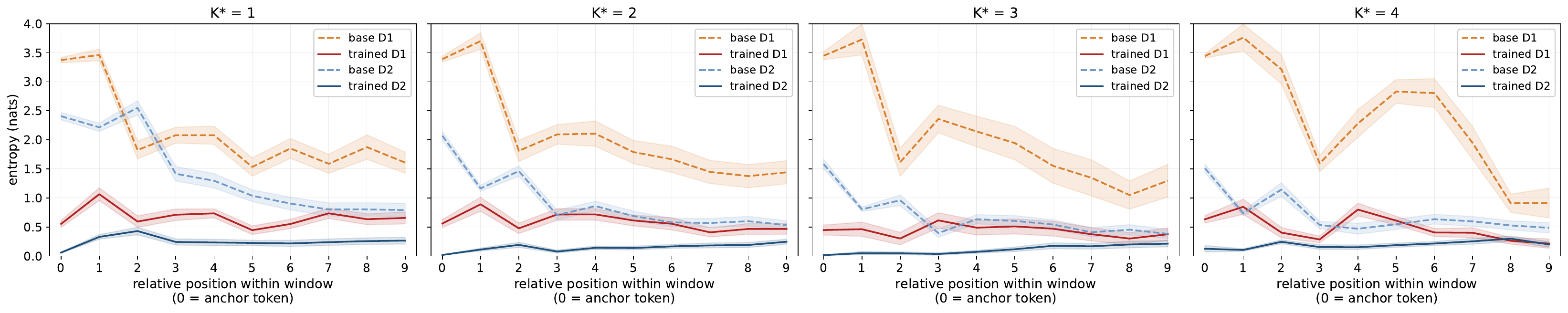}
  \caption{Window-aggregated entropy (mean $\pm$ 95\% CI) at
\texttt{[B]} ($\mathrm{D}_1$) and \texttt{[SEP]} ($\mathrm{D}_2$).
The $\mathrm{D}_1$ gap vanishes at $K^\star\!=\!1$ and grows with
$K^\star$.}
  \vspace{-0.1in}
  \label{fig:agg_entropy}
\end{figure}

We probe whether $\mathcal{L}_{\mathrm{flow}}$ leaves observable signatures in token-level entropy through two diagnostic windows. $\mathrm{D}_1$ spans the 10 tokens after each \texttt{[B]} (hypothesis opening); $\mathrm{D}_2$ spans the 10 tokens after each \texttt{[SEP]} (answer span). The two windows isolate the two loss terms: $\mathcal{L}_{\mathrm{suf}}$ predicts trained $\mathrm{D}_2$ collapses toward zero, while $\mathcal{L}_{\mathrm{sep}}$ predicts trained $\mathrm{D}_1$ retains nontrivial mass over $\mathcal{Z}^\star(x)\setminus\{\hat{z}_{<k}\}$.

Figure~\ref{fig:per_sample_entropy} illustrates both effects on representative samples; Figure~\ref{fig:agg_entropy} confirms them in expectation. On every $K^\star\!>\!1$ panel, trained $\mathrm{D}_2$ lies strictly below the base with disjoint confidence intervals across all 10 positions, while trained $\mathrm{D}_1$ retains boundary mass that decays only gradually. Two regularities sharpen the reading. (i) On $K^\star\!=\!1$, trained and base $\mathrm{D}_1$ coincide: $\mathcal{L}_{\mathrm{sep}}$ admits no active pair, so the absence of a boundary spike reflects the objective by design. (ii) The $\mathrm{D}_1$ gap grows monotonically from $K^\star\!=\!2$ to $K^\star\!=\!4$, contrary to collapse---under separation, each new element of $\mathcal{Z}^\star(x)$ must be held distinct from those already opened, registering its uncertainty at the \texttt{[B]} that opens it.

Read with Table~\ref{tab:main}, this dissociation localizes the effect of $\mathcal{L}_{\mathrm{flow}}$ to the token positions our derivation predicts---evidence that \textsc{FlowEdit} shapes the reasoning trajectory at the boundaries where flows are addressed, rather than producing diffuse gains across the sequence.

\subsection{Ablation Study}
\label{sec:ablation}

Removing either component of $\mathcal{L}_{\mathrm{flow}}$ degrades both metrics, but dropping $\mathcal{L}_{\mathrm{suf}}$ produces a markedly larger drop on EM ($-0.07$) and IR ($-0.05$) than dropping $\mathcal{L}_{\mathrm{sep}}$ ($-0.04$ and $-0.03$), identifying sufficiency as the dominant term. This matches the geometric roles in \S3.2: $\mathcal{L}_{\mathrm{suf}}$ anchors each $e_k$ to $e_{y_k}$ and fixes what each flow encodes, while $\mathcal{L}_{\mathrm{sep}}$ only requires distinct $e_i, e_j$. Without the anchor, separation spreads representations apart in directions that need not correspond to distinct answers, explaining why $\mathcal{L}_{\mathrm{gen}}$ alone matches $\mathcal{L}_{\mathrm{gen}} + \mathcal{L}_{\mathrm{sep}}$ to within seed variance ($0.337$ vs $0.330$ EM): the dual objective is coupled rather than additive.

Replacing $e_a$ with $e_x$ in Eq.~\ref{eq:wij} produces an EM degradation of $-0.07$, comparable to removing $\mathcal{L}_{\mathrm{suf}}$ outright. This is the empirical counterpart of Proposition~1: under $e_x$ the regularizer also penalises the partition-encoding dependence $\Delta_{ij}$ that $A$ contributes beyond $X$, forcing boundary embeddings to forget precisely the structure distinct flows must encode differently. That a mis-conditioned separation term is as costly as undermining sufficiency justifies treating $\epsilon$-sufficiency as load-bearing rather than heuristic. \textsc{FlowEdit} outperforms every ablation on both metrics across all three domains.

 \begin{wraptable}{r}{0.5\linewidth}
  \centering
  \vspace{-10pt}
  \caption{Ablation on Qwen2.5-3B-Instruct.}
  \label{tab:ablation}
  \setlength{\tabcolsep}{3pt}
  \renewcommand{\arraystretch}{1.10}
  \resizebox{\linewidth}{!}{%
  \begin{tabular}{l ccccc}
  \toprule
  \rowcolor{sectionshade}
  \multicolumn{6}{l}{\textbf{Performance (EM)}} \\
  \textbf{Domain}
    & w/o $\mathcal{L}_{\text{flow}}$
    & w/o $\mathcal{L}_{\text{sep}}$
    & w/o $\mathcal{L}_{\text{suf}}$
    & w/o $e_a$
    & \cellcolor{ourshade}\textsc{FlowEdit} \\
  \midrule
  Pure Math   & 0.35 & 0.34 & 0.29 & 0.33 & \cellcolor{ourshade}\textbf{0.39} \\
  Daily       & 0.34 & 0.38 & 0.37 & 0.34 & \cellcolor{ourshade}\textbf{0.42} \\
  Application & 0.32 & 0.35 & 0.33 & 0.31 & \cellcolor{ourshade}\textbf{0.38} \\
  \midrule
  \rowcolor{sectionshade}
  \multicolumn{6}{l}{\textbf{Informativeness (IR)}} \\
  \textbf{Domain}
    & w/o $\mathcal{L}_{\text{flow}}$
    & w/o $\mathcal{L}_{\text{sep}}$
    & w/o $\mathcal{L}_{\text{suf}}$
    & w/o $e_a$
    & \cellcolor{ourshade}\textsc{FlowEdit} \\
  \midrule
  Pure Math   & 0.37 & 0.36 & 0.31 & {0.35} & \cellcolor{ourshade}\textbf{0.39} \\
  Daily       & 0.39 & 0.42 & 0.41 & 0.40 & \cellcolor{ourshade}\textbf{0.45} \\
  Application & 0.38 & 0.39 & 0.40 & 0.39 & \cellcolor{ourshade}\textbf{0.43} \\
  \bottomrule
  \end{tabular}}
  \vspace{-10pt}
  \end{wraptable}

\section{Conclusion}

We presented \textsc{FlowEdit}, a framework that recasts reasoning under conflicting conditions from a detection problem into a flow-regulation problem: rather than asking whether an ill-posed input should be rejected, we ask how a model should keep several valid resolutions distinct as a single autoregressive trace unfolds. The two requirements this entails, namely that each flow be sufficient for its own answer and that distinct flows not collapse into paraphrases of one another, admit a clean expression as conditional mutual information objectives on boundary representations. Across three domains and four branch counts, \textsc{FlowEdit} with Qwen3-4B-Base improves exact-set-match accuracy by $68\%$ and information recovery by $24\%$ over the strongest closed-source baseline, with gains widening as $K^{\star}$ grows. Token-level analysis further reveals that these gains coincide with a structured redistribution of next-token entropy: uncertainty concentrates inside each branch and amplifies at the boundaries between flows, with the boundary signature scaling monotonically with the number of resolutions the problem requires. These results give a representation-level account of why parallel hypothesis maintenance succeeds where sequential elaboration does not. More broadly, this work suggests that the limits of next-token prediction on open-world inputs may be addressable by shaping the internal information geometry of the trace itself. A limitation of the present scope is that \textsc{FlowEdit} regulates flow geometry once branching has begun; the upstream decision of {whether} to branch remains implicit in the policy and are not the objects of the loss, which could be considered for the next step of studying internal reasoning flows.

\bibliography{ref}
\bibliographystyle{unsrtnat}

\newpage
\appendix
\section{Proof of Proposition 1}
\label{app:proof}

We restate the setting before the proof for self-containedness. Let
$X$ denote the random problem statement and $A$ the analysis text generated
from $X$, so that $e_a = h_a(X, A)$ and $e_x = h_x(X)$ are deterministic
functions of their attention contexts. The boundary embeddings $E_1,\dots,E_K$
are produced autoregressively after $A$. We assume the $\epsilon$-sufficiency
bounds
\begin{equation}
I\!\bigl((X, A);\,(E_i, E_j)\,\big|\,e_a\bigr) \;\le\; \epsilon_a,
\qquad
I\!\bigl(X;\,(E_i, E_j)\,\big|\,e_x\bigr) \;\le\; \epsilon_x,
\label{eq:eps_suf}
\end{equation}
and write $I^\star_{ij} := I(E_i; E_j \mid X, A)$ and
$\Delta_{ij} := I(E_i; E_j \mid X) - I^\star_{ij}$.

\subsection*{Two elementary facts}

\textbf{(F1) Determinism.}\ \  If $C = g(U)$ then
$I(\,\cdot\,;\,\cdot\mid U, C) = I(\,\cdot\,;\,\cdot\mid U)$.

\textbf{(F2) Chain rule.}\ \  $I(U; V, W \mid C) = I(U; V \mid C) + I(U; W \mid C, V)$,
all three terms non-negative; this holds for arbitrary conditioning $C$.

\subsection*{Step 1: bound $|I(E_i;E_j\mid e_a) - I^\star_{ij}|$}

Apply (F2) to $I((X,A);(E_i,E_j)\mid e_a)$ in two orderings:
\begin{align}
I\bigl((X,A);(E_i,E_j)\mid e_a\bigr)
&= I\bigl((X,A); E_i \mid e_a\bigr) + I\bigl((X,A); E_j \mid e_a, E_i\bigr) \notag\\
&= I\bigl((X,A); E_j \mid e_a\bigr) + I\bigl((X,A); E_i \mid e_a, E_j\bigr).
\label{eq:step1_chain}
\end{align}
All four summands are non-negative; \eqref{eq:eps_suf} bounds the LHS by
$\epsilon_a$, so each summand is $\le \epsilon_a$. In particular,
\begin{equation}
I\bigl((X,A); E_i \mid e_a\bigr)\le\epsilon_a,
\qquad
I\bigl((X,A); E_i \mid e_a, E_j\bigr)\le\epsilon_a.
\label{eq:step1_star}
\end{equation}

Apply (F2) to $I(E_i; (X,A), E_j \mid e_a)$ in two orderings:
\begin{align}
I(E_i; (X,A), E_j \mid e_a)
&= I(E_i; E_j \mid e_a) + I\bigl(E_i; (X,A) \mid e_a, E_j\bigr) \notag\\
&= I\bigl(E_i; (X,A) \mid e_a\bigr) + I\bigl(E_i; E_j \mid e_a, X, A\bigr).
\label{eq:step1_two_orderings}
\end{align}
Equating the two expansions and rearranging,
\begin{equation}
I(E_i; E_j \mid e_a) - I(E_i; E_j \mid e_a, X, A)
= I\bigl(E_i; (X,A) \mid e_a\bigr) - I\bigl(E_i; (X,A) \mid e_a, E_j\bigr).
\label{eq:step1_dagger}
\end{equation}
By symmetry of mutual information in its two arguments, the two RHS terms in
\eqref{eq:step1_dagger} equal the two quantities bounded in
\eqref{eq:step1_star} respectively, so each lies in $[0, \epsilon_a]$. The
difference of two values in $[0, \epsilon_a]$ lies in $[-\epsilon_a, \epsilon_a]$,
hence
\begin{equation*}
\bigl|\,I(E_i; E_j \mid e_a) - I(E_i; E_j \mid e_a, X, A)\,\bigr| \;\le\; \epsilon_a.
\end{equation*}
Since $e_a = h_a(X, A)$, applying (F1) with $U = (X, A)$ gives
$I(E_i; E_j \mid e_a, X, A) = I(E_i; E_j \mid X, A) = I^\star_{ij}$, so
\begin{equation}
\bigl|\,I(E_i; E_j \mid e_a) - I^\star_{ij}\,\bigr| \;\le\; \epsilon_a.
\label{eq:step1_result}
\end{equation}

\subsection*{Step 2: bound $|I(E_i;E_j\mid e_x) - (I^\star_{ij} + \Delta_{ij})|$}

We repeat Step 1 with the substitution $(X,A)\!\mapsto\!X$,
$e_a\!\mapsto\!e_x$, $\epsilon_a\!\mapsto\!\epsilon_x$. The logical structure
is identical, so we state the substituted intermediate results.

Applying (F2) to $I(X; (E_i, E_j) \mid e_x)$ in two orderings and using
\eqref{eq:eps_suf} yields, in analogy with \eqref{eq:step1_star},
\begin{equation}
I(X; E_i \mid e_x) \le \epsilon_x,
\qquad
I(X; E_i \mid e_x, E_j) \le \epsilon_x.
\label{eq:step2_starstar}
\end{equation}
Applying (F2) to $I(E_i; X, E_j \mid e_x)$ in two orderings and rearranging,
\begin{equation*}
I(E_i; E_j \mid e_x) - I(E_i; E_j \mid e_x, X)
= I(E_i; X \mid e_x) - I(E_i; X \mid e_x, E_j),
\end{equation*}
whose RHS lies in $[-\epsilon_x, \epsilon_x]$ by symmetry of MI and
\eqref{eq:step2_starstar}. Since $e_x = h_x(X)$, (F1) with $U = X$ gives
$I(E_i; E_j \mid e_x, X) = I(E_i; E_j \mid X)$, so
\begin{equation*}
\bigl|\,I(E_i; E_j \mid e_x) - I(E_i; E_j \mid X)\,\bigr| \;\le\; \epsilon_x.
\end{equation*}
By definition $I(E_i; E_j \mid X) = I^\star_{ij} + \Delta_{ij}$, giving
\begin{equation}
\bigl|\,I(E_i; E_j \mid e_x) - (I^\star_{ij} + \Delta_{ij})\,\bigr| \;\le\; \epsilon_x.
\label{eq:step2_result}
\end{equation}

\subsection*{Step 3: combine}

From \eqref{eq:step1_result} and \eqref{eq:step2_result},
$I(E_i;E_j\mid e_a) \le I^\star_{ij} + \epsilon_a$ and
$I(E_i;E_j\mid e_x) \ge I^\star_{ij} + \Delta_{ij} - \epsilon_x$. Subtracting,
\begin{equation}
I(E_i;E_j\mid e_x) - I(E_i;E_j\mid e_a) \;\ge\; \Delta_{ij} - (\epsilon_a + \epsilon_x).
\label{eq:step3_result}
\end{equation}

\subsection*{Extension to arbitrary $\sigma(X)$-measurable $C$}

For any $C = g(X)$ satisfying $I(X; (E_i, E_j) \mid C) \le \epsilon_x$, Step~2
used only (i) the determinism property (F1), which applies because $C$ is a
deterministic function of $X$, and (ii) the $\epsilon_x$-sufficiency bound
with $C$ in place of $e_x$. Both conditions hold for any such $C$, so
$|I(E_i; E_j \mid C) - I(E_i; E_j \mid X)| \le \epsilon_x$ and the lower bound
\eqref{eq:step3_result} carries over with $e_x$ replaced by $C$.
\hfill$\square$

\subsection*{On the sign of $\Delta_{ij}$}

Proposition~1 is stated for arbitrary $\Delta_{ij}$; the claim that
conditioning on $e_a$ rather than $e_x$ \emph{shrinks} cross-flow MI requires
in addition that $\Delta_{ij} \ge 0$. We treat this as a property of the data
construction rather than as a consequence of the graphical structure alone.
The pipeline of Section~4.1 generates each $a^\star$ as a verbalisation of
the latent decomposition that already determines
$\{(z^\star_k, y^\star_k)\}_{k=1}^{K^\star}$ from $x$, so $A$ contributes a
refinement of partition information already implicit in $X$ rather than an
independent source of dependence between branches. Strict positivity,
$\Delta_{ij} > 0$, then corresponds to the multi-flow regime in which $A$
identifies which $z^\star_k$ each branch will encode and is therefore the
regime in which $\mathcal{L}_{\mathrm{sep}}$ is intended to act; the empirical
counterpart is the no-analysis ablation reported in Section~4.4, where
replacing $e_a$ with $e_x$ in $w_{ij}$ degrades EM by $0.07$, comparable to
removing $\mathcal{L}_{\mathrm{suf}}$ outright.

\section{Further Analysis}
\subsection{Sensitivity to the Flow-Regulation Weight \texorpdfstring{$\lambda$}{lambda}}
\label{appn:parameter_sensitivity}

\begin{figure}[H]
  \vspace{-0.1in}
  \centering
  \includegraphics[width=\linewidth]{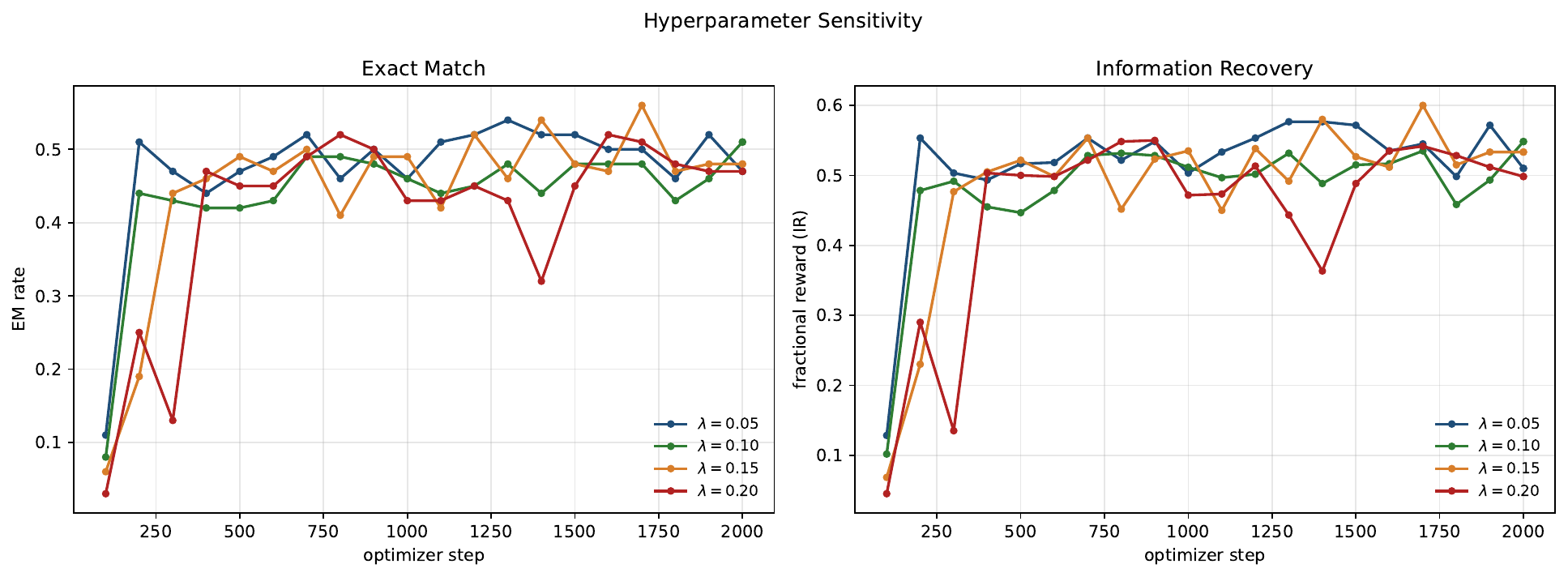}
  \caption{Training dynamics under varying flow-regulation weight
$\lambda \in \{0.05, 0.10, 0.15, 0.20\}$ on Qwen3-4B-Base. EM (left) and
IR (right) are reported every $100$ optimizer steps. Trajectories at
$\lambda \in \{0.05, 0.10\}$ remain confined to a narrow band after
step~$\sim\!500$, whereas $\lambda{=}0.15$ and $\lambda{=}0.20$ exhibit
visible excursions, indicating that stability degrades with $\lambda$.}
  \vspace{-0.1in}
  \label{fig:lambda_dynamics}
\end{figure}

\begin{figure}[H]
  \vspace{-0.1in}
  \centering
  \includegraphics[width=\linewidth]{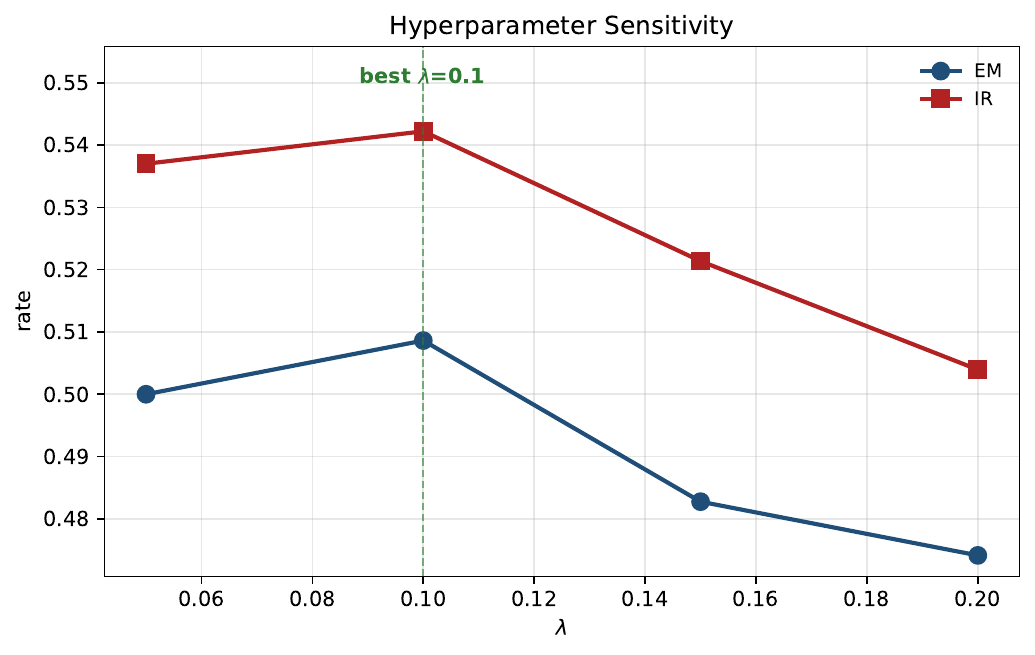}
  \caption{End-of-training EM and IR as a function of the flow-regulation
weight $\lambda$, selected by validation EM. Both metrics peak jointly at
$\lambda{=}0.10$ (EM${=}0.508$, IR${=}0.542$) and decline monotonically on
either side, with steeper degradation above the optimum than below.}
  \vspace{-0.1in}
  \label{fig:lambda_final}
\end{figure}

We sweep $\lambda \in \{0.05, 0.10, 0.15, 0.20\}$ on Qwen3-4B-Base with all
other settings fixed. Figure~\ref{fig:lambda_dynamics} reports per-step
trajectories; Figure~\ref{fig:lambda_final} reports end-of-training values
selected by validation EM.

\textbf{Concave response with interior optimum.} EM and IR peak jointly at
$\lambda{=}0.10$ (EM${=}0.508$, IR${=}0.542$) and decline monotonically on
either side, with the EM--IR gap staying within $[0.030, 0.038]$. The two
metrics co-vary because a single weight scales both
$\mathcal{L}_{\mathrm{suf}}$ and $\mathcal{L}_{\mathrm{sep}}$, which act on
disjoint representational targets and are jointly capacity-bound rather than
antagonistic.

\textbf{Asymmetric degradation.} Raising $\lambda$ from $0.10$ to $0.20$ costs
$0.034$ EM and $0.038$ IR; lowering it to $0.05$ costs only $0.008$ EM and
$0.005$ IR. At large $\lambda$, the geometric constraints on boundary
embeddings overwhelm the token-level gradient, trading branch correctness for
representational separation; at small $\lambda$, $\mathcal{L}_{\mathrm{gen}}$
remains the dominant signal and a directionally correct nudge from
$\mathcal{L}_{\mathrm{flow}}$ recovers most of the gain.

\textbf{Stability degrades with $\lambda$.} All runs reach their plateau by
step~$\sim\!500$. Beyond this, $\lambda \in \{0.05, 0.10\}$ remain in a narrow
band; $\lambda{=}0.15$ shows excursions of $\sim\!0.10$~EM near step~$800$;
and $\lambda{=}0.20$ drops from $\sim\!0.52$ to $\sim\!0.32$ near step~$1400$
before partial recovery. The variance of $\hat{I}_{\mathrm{CLUB}}$ scales with
$\lambda$, and at large weight becomes commensurate with the token-level
signal despite the $[-1,+1]$ clip; $\lambda{=}0.10$ keeps the two losses in
stable equilibrium.

\textbf{Recommendation.} Best-step EM spans $[0.474, 0.508]$ and IR spans
$[0.504, 0.542]$ across the sweep, so FlowEdit is not sensitive to fine
tuning of $\lambda$. We adopt $\lambda{=}0.10$ throughout.

\subsection{Further Reading of the Token-Level Entropy Signature}
\label{app:entropy_further}

Section~\ref{sec:experiments} establishes that $\mathcal{L}_{\mathrm{flow}}$
leaves dissociated marks on the two diagnostic windows. We add one
observation that strengthens the representation-level reading without
introducing additional measurements.

The redistribution induced by $\mathcal{L}_{\mathrm{flow}}$ is not a
uniform damping of next-token uncertainty. Across all $K^\star$ in
Figure~\ref{fig:agg_entropy}, the trained $\mathrm{D}_2$ curve sits at
a near-zero plateau with disjoint confidence bands from base, indicating
that the answer span has become almost deterministic once a hypothesis
is committed. The same trained model retains visibly higher
$\mathrm{D}_1$ mass at the branch-opening anchor, so the residual
uncertainty the policy preserves is concentrated at the tokens that
designate which hypothesis a new flow will encode rather than at the
tokens that elaborate it. The two windows therefore describe a
redistribution of next-token uncertainty rather than its global
suppression: the trace becomes more decisive within each branch
precisely at the positions where commitment to a particular flow has
already been made.

\section{Implementation Details}
\label{sec:appn_implementation}
This appendix consolidates the training and evaluation details required to
reproduce the results in Section~\ref{sec:experiments}: dataset splits, the
joint training procedure, optimizer configuration, decoding settings for evaluation, and the compute used to obtain experiment data. 

\subsection{Data Splits}
\label{ssec:appx_data}

The 5{,}000 ill-posed problems described in Section~\ref{sec:experiments} are
partitioned once into a training and a held-out validation split, fixed across
all runs. We use a 90/10 split stratified jointly over branch count
$K^{\star}\!\in\!\{1,2,3,4\}$ and the three domain categories (Pure Math,
Daily, Application), so the marginal distribution of $K^{\star}$ and of domain
in validation matches that in training; this prevents differences in
$K^{\star}$ frequency from being absorbed into the metric.

\subsection{Backbones and Adapter Configuration}

\textsc{FlowEdit} is instantiated on \textsc{Qwen3-4B-Base} and
\textsc{Qwen2.5-3B-Instruct} \cite{yang2025qwen3}. We attach a LoRA
\cite{hu2022lora} adapter to the four attention projections
$\{q,k,v,o\}$ of every transformer block (rank $r{=}16$,
$\alpha{=}32$, dropout~0.05); MLP projections are kept frozen. The six
structural tokens $\{$\texttt{[ANA]}, \texttt{[/ANA]}, \texttt{[B]},
\texttt{[SEP]}, \texttt{[/B]}, \texttt{[END]}$\}$ extend the tokenizer
vocabulary; their input and output embedding rows are added to the trainable
parameter set so the policy can both consume and emit them. For
\textsc{Qwen2.5-3B-Instruct}, which uses tied input/output embeddings, this
means a single shared row per special token is updated. All training is in
\texttt{bfloat16} with \texttt{flash\_attention\_2}; activations are
recomputed via gradient checkpointing.

\subsection{Training Procedure}
\label{ssec:appx_training}

Both backbones are trained directly from the base checkpoint by jointly
optimizing $\mathcal{L}_{\mathrm{gen}} + \lambda\,\mathcal{L}_{\mathrm{flow}}$
for 2{,}000 optimizer steps, with $\lambda$ fixed at $0.10$ throughout
training. The token-level term $\mathcal{L}_{\mathrm{gen}}$ in
Eq.~\ref{eq:lgen} and the flow-regulation term $\mathcal{L}_{\mathrm{flow}} =
\mathcal{L}_{\mathrm{suf}} + \mathcal{L}_{\mathrm{sep}}$ in Eq.~\ref{eq:lflow}
are computed on every mini-batch and back-propagated through the same set of
trainable parameters (LoRA adapter $+$ special-token embedding rows). No
warm-start, distillation, or staged ramp on $\lambda$ is used; the joint
objective is active from the first optimizer step.

The CLUB network $q_{\psi}$ is a two-layer diagonal-Gaussian MLP with hidden
size $256$ and dropout~0.1, predicting the mean and log-variance of $e_j$ given
the conditioning $(e_i, w_{ij})$ in Eq.~\ref{eq:wij}. To keep the upper-bound
property of $\widehat{I}_{\mathrm{CLUB}}$ approximately tight, $q_{\psi}$ is
updated three times per outer optimizer step using its own AdamW optimizer
(see Table~\ref{tab:hparams}) before the policy step that consumes its score.
The InfoNCE scoring head $f_{\phi}$ has the same hidden size and dropout, is
updated jointly with the policy, and uses temperature $\tau{=}0.07$. The per-pair CLUB term is clipped to
$[-1, +1]$ before averaging, as stated in Eq.~\ref{eq:lsepterm}, to bound the
contribution of any single mini-batch pair to a fixed nats budget.

All entries in
Table~\ref{tab:hparams} were either fixed at their HuggingFace Trainer
defaults or adopted unchanged from the LoRA fine-tuning recipe shipped with
the Qwen 2.5 / 3 release \cite{yang2025qwen3}, and reused across both
backbones and every ablation arm without per-arm tuning. We use random seed
$42$ for primary runs and report mean $\pm$ one standard deviation across
three independent seeds in Fig.~\ref{fig:training_dynamics} and
Tab.~\ref{tab:main}.

\begin{table}[t]
\centering
\small
\setlength{\tabcolsep}{6pt}
\renewcommand{\arraystretch}{1.05}
\begin{tabular}{l l l}
\toprule
\textbf{Group} & \textbf{Hyperparameter} & \textbf{Value} \\
\midrule
\multirow{7}{*}{Policy optimizer}
 & Optimizer                          & AdamW \\
 & Learning rate                      & $5\!\times\!10^{-5}$ \\
 & LR schedule                        & Linear warm-up + cosine decay \\
 & Warm-up ratio                      & 0.05 \\
 & Weight decay                       & 0.01 \\
 & Adam $(\beta_1, \beta_2, \epsilon)$ & $(0.9,\,0.999,\,1\!\times\!10^{-8})$ \\
 & Gradient clip ($\ell_2$ norm)      & 1.0 \\
\midrule
\multirow{4}{*}{CLUB optimizer ($q_\psi$)}
 & Optimizer                          & AdamW \\
 & Learning rate                      & $5\!\times\!10^{-4}$ \\
 & Inner updates per outer step       & 3 \\
 & Adam $(\beta_1, \beta_2, \epsilon)$ & $(0.9,\,0.999,\,1\!\times\!10^{-8})$ \\
\midrule
\multirow{4}{*}{Batch}
 & Per-device batch size              & 2 \\
 & Gradient accumulation              & 4 (4B, 2-GPU) / 8 (3B, 1-GPU) \\
 & Effective batch (problems / step)  & 16 \\
 & Maximum sequence length            & 2048 \\
\midrule
\multirow{2}{*}{Duration}
 & Total optimizer steps              & 2{,}000 \\
 & Mixed precision                    & bfloat16 \\
\bottomrule
\end{tabular}
\caption{Optimizer and step-size settings used to train \textsc{FlowEdit}.
The same configuration is used for both \textsc{Qwen3-4B-Base} and
\textsc{Qwen2.5-3B-Instruct} except for the gradient-accumulation factor,
which is adjusted to keep the effective batch size constant across the
two-GPU (4B) and one-GPU (3B) configurations.}
\label{tab:hparams}
\end{table}

\subsection{Decoding and Evaluation Settings}

For every evaluation point reported in the paper, including the training
dynamics curves in Fig.~\ref{fig:training_dynamics}, we decode the validation
split with nucleus sampling at $T{=}0.7$, $p{=}0.9$, with a hard cap of
$1024$ new tokens for \textsc{Qwen3-4B-Base} and $1500$ for
\textsc{Qwen2.5-3B-Instruct}, and \texttt{[END]} as the stop string. Decoding
is done with the LoRA adapter merged into the base model. The branches
$\hat{Y}(x)$ are recovered from the decoded string by parsing the
\texttt{[B] $\cdot$ [SEP] $\cdot$ [/B]} blocks; matches against $Y^{\star}(x)$
use numeric tolerance $10^{-3}$ for floating-point answers and a SymPy-based
canonicalization for symbolic ones, identically across
\textsc{FlowEdit} and the prompted baselines. We average each
$(K^{\star}, \text{domain})$ cell over all examples that fall in it, and then
average those cell values to form the per-domain and overall
columns of Table~\ref{tab:main}. Variance across the three independent seeds
is reported as the shaded band in Fig.~\ref{fig:training_dynamics} (one
standard deviation).

\subsection{Compute and Wall-Clock Time}
\label{ssec:appx_compute}

All experiments were run on NVIDIA H100~80\,GB GPUs through PyTorch
$\geq$2.3 with HuggingFace Transformers $\geq$4.46 and PEFT $\geq$0.14.
\textsc{Qwen3-4B-Base} was trained on two GPUs with DDP and mixed-precision
\texttt{bfloat16}; \textsc{Qwen2.5-3B-Instruct} fits on a single H100 and
was trained without DDP. Joint training of the
$\mathcal{L}_{\mathrm{gen}} + \lambda\,\mathcal{L}_{\mathrm{flow}}$ objective
for 2{,}000 optimizer steps takes approximately 15 hours of wall-clock time
per backbone per seed.

Including the three independent seeds reported in
Fig.~\ref{fig:training_dynamics} and Tab.~\ref{tab:main}, the four ablation
arms in Tab.~\ref{tab:ablation} (\,w/o $\mathcal{L}_{\mathrm{flow}}$, w/o
$\mathcal{L}_{\mathrm{sep}}$, w/o $\mathcal{L}_{\mathrm{suf}}$, prompt-only
conditioning $e_x$\,), and the four-point $\lambda$ sweep in
Appendix~\ref{appn:parameter_sensitivity}, the total compute spent on the
reported numbers is on the order of several hundred H100-GPU-hours.
Closed-source baselines (Claude Haiku 4.5, GPT-5, Gemini 2.5 Pro) were
evaluated through their public APIs and contributed only inference cost
(no training).

\section{Dataset Details}

\subsection{Dataset Construction Pipeline}
\label{app:dataset_construction}

The dataset is produced by a four-stage pipeline that fixes the conflict 
structure $K^\star$ before any natural language is generated, then realizes 
a problem statement around it. All four stages are implemented as 
independent calls to a single frontier LLM; no other model is involved, 
and the conflict skeleton produced in Stage~1 is preserved verbatim 
through subsequent stages so that $K^\star$ is determined by construction 
rather than recovered post hoc.

\paragraph{Stage 1: Conflict-structure design.} 
For each instance we sample $K^\star$ from a target distribution over 
$\{1,2,3,4\}$, a sub-domain from a fixed pool that groups into the three 
high-level domains (Pure Math, Daily, 
Application), and a difficulty tier from $\{$middle school, high school, 
early college$\}$. The model is prompted to design the conflict skeleton 
before any prose: it nominates the conflicting quantity, lists its 
$K^\star$ candidate values, and specifies a multi-step computation chain 
that propagates the conflicting quantity to the final answer. The chain 
is required to contain enough computation steps that branches are 
distinguished by genuine derivation rather than by surface substitution. 
A self-check at the end of the stage asserts that all 
$K^\star$ branch answers are distinct and that no condition is entangled 
across branches; failures are discarded.

\paragraph{Stage 2: Surface realization.} 
A second call rewrites the Stage-1 skeleton into the user-facing problem 
$x$, identifies the contradicting conditions, writes the shared analysis 
prose $a$, and emits a list of $K^\star$ branch records, each specifying 
the trusted and discarded condition subset and the well-posed sub-problem 
the branch solves. No numeric answer is produced at this stage.

\paragraph{Stage 3: Per-branch solving.} 
Each branch's sub-problem is dispatched as an independent solver call 
that returns a derivation ending in a boxed final value. The narrative 
may use only the trusted conditions of that branch, and every arithmetic 
step is shown explicitly to support the verification in Stage~4.

\paragraph{Stage 4: Verification, filtering, and spot-check.} 
A fourth call verifies each branch against five criteria: assumption 
coherence with the trusted condition subset, step-by-step logical 
validity, arithmetic correctness recomputed by an independent symbolic 
evaluator, answer--narrative agreement, and overall narrative quality. 
The arithmetic check is performed outside the LLM loop so that 
verification of numeric correctness does not depend on the same model 
that produced the derivation. A branch is retained only if all five 
criteria pass. We then apply exact and near-duplicate removal on problem 
text, drop instances whose branches collide on the boxed value (which 
would collapse $|\mathcal{Y}^\star(x)|$ below $K^\star$), and discard 
samples with empty or non-informative final answers. A subsequent human 
spot-check on a stratified random sample is conducted to confirm 
mathematical correctness, conflict structure, and analysis fidelity, 
with failed instances returned to Stage~2 for re-realization rather 
than silently dropped, to avoid biasing the difficulty distribution.

\paragraph{Final composition.} 
The released dataset contains 5{,}000 instances spanning the three 
domains and the four values of $K^\star$. The marginal distribution of 
$K^\star$ is skewed toward smaller branch counts, reflecting that 
$K^\star{=}4$ instances require two independent binary conflicts to 
co-occur in a single coherent problem statement and are correspondingly 
rarer; we explicitly include $K^\star{=}1$ instances so that the model 
cannot exploit a fixed enumeration prior. 
Because problems are synthesized from conflict-structure templates 
rather than collected from existing sources, contamination with public 
math benchmarks is not expected by construction.

 To illustrate the structure described in Section~\ref{sec:experiments}, we
  include one representative training instance for each value of
  $K^{\star}\!\in\!\{1,2,3,4\}$. Every instance is stored as a JSON object
  with the same schema---\texttt{problem}, \texttt{analysis}, and a list of
  \texttt{branches}, each carrying a \texttt{hypothesis} and the
  corresponding \texttt{answer}---together with the bookkeeping fields
  \texttt{K}, \texttt{id}, \texttt{domain}, and \texttt{difficulty}.
  Listings~\ref{lst:k1}--\ref{lst:k4} step through the four cases: the
  degenerate $K^{\star}=1$ instance where multi-source data verifies as
  consistent and the framework reduces to standard reasoning; a binary
  $K^{\star}=2$ conflict between two authoritative specifications; a
  $K^{\star}=3$ instance with three mutually exclusive candidate values for
  a single quantity; and a $K^{\star}=4$ instance whose two independent
  binary disagreements yield four maximal self-consistent combinations.
  Long \texttt{analysis} and \texttt{answer} bodies are abbreviated with
  ``\texttt{[...]}'' for readability; the underlying records are stored
  verbatim and used unmodified during training.
\label{appn:dataset}

\begin{lstlisting}[language=json,caption={Dataset example with $K^{\star}=1$ (well-posed degenerate case: the multi-source data \emph{appears} to conflict, but arithmetic verification shows the conditions are jointly consistent, so the framework collapses to standard single-trace reasoning). Domain: \emph{combinatorics / counting}; difficulty: middle school.},label={lst:k1}]
{
  "id": "sample_2368",
  "K": 1,
  "domain": "combinatorics / counting",
  "difficulty": "middle school",
  "problem": "A jar contains red marbles and blue marbles. There are 5 blue marbles in the jar. The red marbles were added in three separate batches: the first batch had 8 red marbles, the second batch had 4 additional red marbles (making 12 total red marbles), and the third batch had 3 more red marbles (making 15 total red marbles). If you randomly select 3 marbles from the jar, how many ways can you select them such that at least one marble is blue?",
  "analysis": "I need to first establish the total composition of marbles in the jar. The problem states there are 5 blue marbles, and describes the red marbles being added in three batches. Let me verify the arithmetic: the first batch contributes 8 red marbles, the second batch adds 4 more bringing the total to 12, and the third batch adds 3 more bringing the final total to 15. Checking this sequence: 8 + 4 = 12 \\checkmark, and 12 + 3 = 15 \\checkmark. The arithmetic is consistent throughout, so the jar contains 5 blue and 15 red marbles for a total of 20. The problem is well-posed and reduces to a standard combinatorics calculation.",
  "branches": [
    {
      "hypothesis": "All given conditions are correct and consistent. The jar contains 5 blue marbles and 15 red marbles (20 total marbles).",
      "answer": "Using complementary counting, the number of ways to pick 3 marbles with at least one blue equals C(20,3) - C(15,3) = 1140 - 455 = 685. The answer is \\boxed{685}."
    }
  ]
}
\end{lstlisting}

\begin{lstlisting}[language=json,caption={Dataset example with $K^{\star}=2$ (binary conflict between architectural and regulatory specifications). Domain: \emph{construction / area \& materials}; difficulty: early college.},label={lst:k2}]
{
  "id": "sample_4101",
  "K": 2,
  "domain": "construction / area and materials",
  "difficulty": "early college",
  "problem": "A construction company is building a rectangular foundation for a warehouse. The foundation has a length of 20 meters and requires concrete to be poured to a depth of 0.6 meters. The concrete costs $300 per cubic meter. According to the architectural plans, the foundation width should be 60% of the length. However, the site supervisor notes that local building codes require the width to be at least 15 meters for this type of structure. What is the total cost of the concrete needed for this foundation?",
  "analysis": "I need to calculate the volume of concrete required and multiply by the cost per cubic meter. [...] The architectural plans call for a width of 0.6 \\times 20 = 12 meters. However, local building codes require a minimum width of 15 meters. This creates a direct conflict: the architectural width of 12 m is less than the legally required minimum of 15 m. Both requirements cannot be satisfied simultaneously, so I must determine which specification takes precedence to proceed with the cost calculation.",
  "branches": [
    {
      "hypothesis": "The architectural plans are authoritative and the building code requirement is either outdated, misapplied, or will be addressed through a variance process",
      "answer": "Width = 0.6 \\times 20 = 12 m. Volume = 20 \\times 12 \\times 0.6 = 144 m^3. Cost = 144 \\times 300 = $43,200. The answer is \\boxed{$43,200}."
    },
    {
      "hypothesis": "The building code requirement is legally binding and takes precedence over the architectural specification, requiring a design modification",
      "answer": "Width = 15 m (code minimum). Volume = 20 \\times 15 \\times 0.6 = 180 m^3. Cost = 180 \\times 300 = $54,000. The answer is \\boxed{$54,000}."
    }
  ]
}
\end{lstlisting}

\begin{lstlisting}[language=json,caption={Dataset example with $K^{\star}=3$ (three mutually exclusive choices among candidate principals). Domain: \emph{finance / interest \& investment}; difficulty: middle school.},label={lst:k3}]
{
  "id": "sample_0898",
  "K": 3,
  "domain": "finance / interest and investment",
  "difficulty": "middle school",
  "problem": "Sarah wants to invest money in a savings account that pays 8% simple interest per year. After 2 years, she will receive a $20 bonus from the bank for being a loyal customer. However, there will be a $20 processing fee when she withdraws her money. Sarah mentions three different amounts she might invest: she could use money from her birthday gifts totaling $200, or she could combine that with her allowance savings of $100 for a total of $300, or she could use her part-time job earnings of $250. How much money will Sarah have after the 2-year investment period, including the bonus and after paying the fee?",
  "analysis": "I need to determine Sarah's final amount after a 2-year investment with 8% simple interest, plus a $20 bonus, minus a $20 processing fee. [...] The problem presents three different possible investment amounts: $200, $300, or $250. The question asks for 'how much money will Sarah have' as if expecting a single answer, but the final amount depends entirely on which initial investment amount she chooses, yet no single choice is specified [...] I cannot proceed with a calculation until I determine which investment scenario to use.",
  "branches": [
    {
      "hypothesis": "Sarah chooses to invest the $200 from her birthday gifts, treating the other amounts as merely considered alternatives",
      "answer": "Interest = 200 \\times 0.08 \\times 2 = $32. Balance = 200 + 32 + 20 - 20 = $232. The answer is \\boxed{$232}."
    },
    {
      "hypothesis": "Sarah chooses to invest the combined $300 from birthday gifts and allowance savings, treating the other amounts as merely considered alternatives",
      "answer": "Interest = 300 \\times 0.08 \\times 2 = $48. Balance = 300 + 48 + 20 - 20 = $348. The answer is \\boxed{$348}."
    },
    {
      "hypothesis": "Sarah chooses to invest the $250 from her part-time job earnings, treating the other amounts as merely considered alternatives",
      "answer": "Interest = 250 \\times 0.08 \\times 2 = $40. Balance = 250 + 40 + 20 - 20 = $290. The answer is \\boxed{$290}."
    }
  ]
}
\end{lstlisting}

\begin{lstlisting}[language=json,caption={Dataset example with $K^{\star}=4$ (two independent binary conflicts---price and discount---producing four maximal self-consistent combinations, each with a distinct final amount). Domain: \emph{shopping / pricing \& discounts}; difficulty: middle school.},label={lst:k4}]
{
  "id": "sample_3941",
  "K": 4,
  "domain": "shopping / pricing and discounts",
  "difficulty": "middle school",
  "problem": "Sarah is buying a backpack and a water bottle. The water bottle costs $12. She has conflicting information about the backpack: one tag shows $40 while another shows $50. The store clerk mentions there's a discount on backpacks, but Sarah heard different amounts - either 20% off or 30% off. After the discount is applied to the backpack, she needs to pay 8% sales tax on her total purchase (backpack + water bottle). Finally, she can use a $5 store credit she earned from previous purchases. How much will Sarah pay in total?",
  "analysis": "I need to calculate Sarah's total cost by determining the discounted backpack price, adding the water bottle cost, applying sales tax, and subtracting the store credit. [...] The backpack has two different price tags showing $40 and $50, which cannot both be correct simultaneously, and the discount rate is reported as both 20% and 30% from different sources. The price disagreement and the discount disagreement are independent: each yields its own pair of self-consistent values, so combining them gives four maximal self-consistent hypotheses, each producing a different final amount.",
  "branches": [
    {
      "hypothesis": "The $40 price tag is correct and the 20% discount applies, while the $50 tag and 30% discount information are errors",
      "answer": "Discounted backpack: 40 \\times 0.80 = $32. Subtotal: 32 + 12 = $44. With 8% tax: 44 \\times 1.08 = $47.52. After $5 credit: $42.52. The answer is \\boxed{$42.52}."
    },
    {
      "hypothesis": "The $40 price tag is correct and the 30% discount applies, while the $50 tag and 20% discount information are errors",
      "answer": "Discounted backpack: 40 \\times 0.70 = $28. Subtotal: 28 + 12 = $40. With 8% tax: 40 \\times 1.08 = $43.20. After $5 credit: $38.20. The answer is \\boxed{$38.20}."
    },
    {
      "hypothesis": "The $50 price tag is correct and the 20% discount applies, while the $40 tag and 30% discount information are errors",
      "answer": "Discounted backpack: 50 \\times 0.80 = $40. Subtotal: 40 + 12 = $52. With 8% tax: 52 \\times 1.08 = $56.16. After $5 credit: $51.16. The answer is \\boxed{$51.16}."
    },
    {
      "hypothesis": "The $50 price tag is correct and the 30% discount applies, while the $40 tag and 20% discount information are errors",
      "answer": "Discounted backpack: 50 \\times 0.70 = $35. Subtotal: 35 + 12 = $47. With 8% tax: 47 \\times 1.08 = $50.76. After $5 credit: $45.76. The answer is \\boxed{$45.76}."
    }
  ]
}
\end{lstlisting}

\section{Prompts}
\label{appn:prompts}

We evaluate models with two prompt strategies that bracket the spectrum of
ill-posedness disclosure: a \emph{blind} prompt that does not mention multiple
interpretations (\ref{fig:prompt_blind}), and an \emph{ill-posed--aware}
prompt that explicitly tells the model the problem admits multiple valid
solutions (\ref{fig:prompt_illposed}). 

\begin{figure}[H]
    \begin{definedbox}{(ill-posed--aware)}
    \textbf{System message:}
    \begin{verbatim}
You are a careful problem solver. The problems you receive
are ILL-POSED -- they typically contain conflicting or
ambiguous conditions that admit multiple valid
interpretations. Each problem has K valid solutions, where K
depends on how many independent contradictions exist. Your
job is to identify each valid interpretation and solve the
problem under each.
    \end{verbatim}

    \textbf{User message:}
    \begin{verbatim}
Solve the following ill-posed problem. The problem has
multiple valid solutions depending on which subset of
conditions you trust. Identify ALL valid interpretations and
provide the answer under each.

Respond with a single JSON object and nothing else:
{
  "analysis": "Identify the specific conflicts or
               ambiguities, then list all possible
               interpretations.",
  "branches": [
    {"hypothesis": "Which conditions are trusted in this
                    interpretation (and which are discarded)",
     "answer": "Final answer under this interpretation
                (number or expression)"}
  ]
}

Critical rules:
- The number of branches MUST equal the number of distinct
  valid interpretations that genuinely arise from
  contradictions in the problem.
- If the problem is actually well-posed (no real
  contradictions), output exactly one branch.
- Do NOT invent speculative branches just to enumerate; only
  include an interpretation if it follows from a real
  conflict between conditions.
- Each branch's answer must be the result computed under
  that branch's hypothesis.

Problem:
{problem}
    \end{verbatim}
    \end{definedbox}
    \caption{Ill-posed--aware prompt. The model is told the input is ill-posed
    and asked to enumerate all valid interpretations.}
    \label{fig:prompt_illposed}
\end{figure}

\begin{figure}[H]
    \begin{definedbox}{(blind)}
    \textbf{System message:}
    \begin{verbatim}
You are a careful problem solver. Read the problem, think step
by step, then give your final answer.
    \end{verbatim}

    \textbf{User message:}
    \begin{verbatim}
Solve the following problem.

Respond with a single JSON object and nothing else. The JSON
must have exactly these fields:
- "reasoning": your step-by-step reasoning as a string
- "answer": your final answer as a string

Problem:
{problem}
    \end{verbatim}
    \end{definedbox}
    \caption{Blind prompt. The model is not told that problems may be ill-posed.}
    \label{fig:prompt_blind}
\end{figure}

\end{document}